\documentclass[review]{elsarticle}

\usepackage{lineno,hyperref}
\modulolinenumbers[5]

\journal{International Journal of Approximate Reasoning}

\usepackage[toc,page]{appendix} 
\usepackage{lineno}  % COMMENT TO REMOVE LINE NUMBERS
\usepackage{relsize}
\usepackage{caption}
\usepackage{subcaption}
\usepackage{hyperref}
\usepackage{tikz}
\usepackage{cancel}
\usepackage{amsmath, amssymb}
\usepackage{multirow}
\usepackage{ulem}
\usepackage[ruled]{algorithm}
\usepackage{algorithmic}
\usepackage{multirow}
\usepackage{hyperref}
\definecolor{blue}{HTML}{1D92D1}
\definecolor{red}{HTML}{FA0004}
\definecolor{green}{HTML}{00A603}
\definecolor{orange}{rgb}{1,0.5,0}

\newcommand{\NPERP}{\mbox{\ensuremath{\not\perp}}}
\newcommand{\indep}[2]{\ensuremath{#1 \perp #2}}
\newcommand{\dep}[2]{\ensuremath{#1 \NPERP #2}}

\newcommand{\ci}[3]{\ensuremath{\langle\indep{#1}{#2} \arrowvert #3}\rangle}
\newcommand{\cd}[3]{\ensuremath{\langle\dep{#1}{#2} | #3}\rangle}
\newcommand{\eq}[1]{(\ref{#1})}

\newcommand{\blanket}[1]{\ensuremath{B^{#1}}}

\newcommand{\closure}{\ensuremath{\mathcal{C}}}

\usepackage{eqparbox}

\begin{document}

\begin{frontmatter}

\title{Blankets Joint Posterior score for learning Markov network structures}

% \author{Federico Schl\"uter \and Yanela Strappa \and Facundo Bromberg }
% \address{DHARMa Lab, Dept of Information Systems.
% Facultad Regional Mendoza, Universidad Tecnol\'ogica Nacional, 
% Mendoza, Argentina. Tel.: +54-261-5240066}
\author[utnaddress]{Federico Schl\"uter}
\cortext[mycorrespondingauthor]{Corresponding author}
\ead{federico.schluter@frm.utn.edu.ar}
\author[utnaddress]{Yanela Strappa}
\author[sinciAddress]{Diego Milone}
\author[utnaddress]{Facundo Bromberg}

\address[utnaddress]{DHARMa Lab, Dept of Information Systems.
Facultad Regional Mendoza, Universidad Tecnol\'ogica Nacional, 
Mendoza, Argentina. Tel.: +54-261-5240066}
\address[sinciAddress]{Research Institute for Signals, Systems and Computational Intelligence, sinc(i),
FICH-UNL/CONICET
Santa Fe, Argentina.}

\begin{abstract}
Markov networks are extensively used to model 
complex sequential, spatial, and relational interactions 
in a wide range of fields. 
By learning the structure of independences of a domain,
more accurate joint probability distributions can be obtained for inference tasks
or, more directly, for interpreting the most significant relations among the variables.
Recently, several researchers have investigated techniques for automatically learning the structure
from data by obtaining the probabilistic maximum-a-posteriori structure given the available data.
However, all the approximations proposed decompose the posterior of the whole structure
into local sub-problems, by assuming that the posteriors of 
the Markov blankets of all the variables
are mutually independent. 
In this work, we propose a scoring function for relaxing such assumption. 
The {\it Blankets Joint Posterior\/} score computes the 
joint posterior of structures as a joint distribution of the collection of its Markov blankets.
Essentially, the whole posterior is obtained by computing the posterior of the blanket of each variable
as a conditional distribution that takes into account information from other blankets in the network. 
We show in our experimental results that the proposed approximation 
can improve the sample complexity of state-of-the-art scores 
when learning complex networks, where the independence assumption between blanket variables is clearly incorrect. 
\end{abstract}


\end{frontmatter}

% \linenumbers
\section{Introduction \label{sec:intro}}
A Markov network (MN) is  a popular probabilistic graphical model that efficiently 
encodes the joint probability distribution for a set of random variables of a specific domain \cite{pearl88,LAURITZEN82,koller09}.
MNs usually represent probability distributions by using two interdependent components: 
an independence structure, 
and a set of numerical parameters over the structure.
The first is a qualitative component that represents
structural information about a problem domain in the form of conditional independence relationships
between variables.  
The numerical parameters are a quantitative component that represents the strength of 
the dependences in the structure. 
There is a large list of applications of MNs
in a wide range of fields, such as 
computer vision and image analysis \cite{Li2006,hwang2015markov,peng2016n},
computational biology \cite{li2015gene},
biomedicine \cite{schmidts08,wan2015package},
and evolutionary computation \cite{larranagalozano2002,shakya2012markovianity}, among many others.
For some of these applications, the model
can be constructed manually by human experts, but in many other problems
this can become unfeasible, mainly due to the dimensionality of the problem.

Learning the model from data consists of two interdependent problems:
learning the structure; and given the structure,
learning its parameters. This work focuses on the task of learning the structure, 
which is useful for a variety of tasks.
The structures learned may be used to construct accurate models for inference
tasks (such as the estimation of marginal and conditional probabilities) \cite{lowd14a,van2012markov,davis2010bottom}, 
and may also be interesting per se,
since they can be used as interpretable models 
that show the most significant interactions of a domain \cite{LeealNIPS06,haaren2013,claeskens2015,nyman2014context,pensar2016marginal}.
The first scenario is known in practice as the density estimation goal of learning,
and the second one is known as the knowledge discovery goal of learning [Chapter~16~\cite{koller09}].
% This work focuses on the second class of algorithms.

An interesting approach to MN structure learning is to use 
constraint-based (also known as independence-based) algorithms \cite{Spirtes00,brombergmargaritis09b,Aliferis2010,schluter2012survey}. 
Such algorithms proceed by performing statistical independence tests on data, 
and discard all structures inconsistent with the tests. This is 
an efficient approach, and it is correct under the assumption that the distribution can be represented by a graph,
and that the tests are reliable. However, the algorithms that follow this approach are quite sensitive to errors in the tests,
which may be unreliable for large conditioning sets \cite{Spirtes00,koller09}.
A second approach to MN structure learning is to use score-based algorithms \cite{PietraPL97,MCCALLUM03,LeealNIPS06,ganapathi2008}.
Such algorithms formulate the problem as an optimization,
combining a strategy for searching through the space of possible structures 
with a scoring function measuring the fitness of each structure to the data. 
The structure learned is the one that achieves the highest score.

It is important to mention that both constraint-based and score-based approaches
have been originally motivated by distinct learning goals.
According to the existing literature \cite{koller09}, 
constraint-based methods are generally designed for the
knowledge-discovery goal of learning \cite{Aliferis2010,brombergmargaritis09b},
and their quality is often measured in terms of the correctness of the structure learned (structural errors).
In contrast, most score-based approaches have been designed for the density estimation goal
of learning \cite{lowd14a,van2012markov,davis2010bottom},
and they are in general evaluated in terms of inference accuracy.
For this reason, score-based algorithms
often work by considering the whole MN at once during the search,
interleaving the parameter learning step.
This makes them more accurate for inference tasks.
However, since learning the parameters is known to be NP-hard for MNs \cite{BARAHONA82},
it has a negative effect on their scalability.

Recently, there has been a surge of interest towards efficient methods based on a 
strategy that follows a score-based approach, but with the knowledge discovery goal in mind.
Basically, an undirected graph structure is learned by 
obtaining the probabilistic maximum-a-posteriori structure given the available data \cite{schluter2014ibmap,pensar2016marginal}.
This hybrid strategy achieves scalability, as well as reliable performance.
Such contributions consist in the design of efficient scoring functions
for MN structures, expressing the problem formally as follows:
given a complete training data set $D$, find an undirected graph $G^{\star}$ such that
\begin{equation} \label{eq:maxG} 
 G^{\star} = \arg\max_{G \in \mathcal{G}}{\Pr(G | D)},
\end{equation}		
where $\Pr(G | D)$ is the posterior probability of a structure given $D$, 
and $\mathcal{G}$ is the familiy of all the possible undirected graphs
for the domain size.
This class of algorithms has been shown to outperform constraint-based algorithms
in the quality of the learned structures,
with equivalent computational complexities. 
The method proposed in this paper follows this approach.

% The Bayesian score can compute the posterior of structures given data, but it 
% only guaranties correctness in distributions where the independence structure is a chordal graph.
% This is an important limitation, because the percentage of chordal networks 
% tends to zero when the size of the domain grow,
% and in the practice it is not possible to assume that the structure would be chordal. 
% The current state-of-the art methods propose approximations that not assume chordality, 
% but still do some approximations. 
% One drawback of the above techniques is that they assume
% that each node has the same amount of L1 regularization;
% this constant is often set by cross validation.
% 

Since there are no feasible exact methods for computing the posterior of MN structures, 
different approximations have been proposed.
An important assumption commonly made by the current state-of-the-art methods
is to suppose that the posterior 
of the structure is decomposable \cite{frydenberg1989decomposition,dawid1993hyper,koller09,schluter2014ibmap,pensar2016marginal}. 
It means that the whole posterior can be computed as a product 
of the posteriors of the Markov blankets that compose the structure,
which are smaller posteriors that can be computed independently.
In fact, this is a good approximation that improves the efficiency of search.
The research line of this work aims at designing a better approximation to the posterior, 
by relaxing such independence assumption. 
% ....in which situations will this learning process perform better?
% and which situations are expected to improve the learning process by using such improved scoring function? 
For this, the contribution of this work is the \textit{Blankets Joint Posterior} (BJP),
a scoring function that poses $\Pr(G | D)$
as the joint posterior probability of the Markov blankets of $G$.
This is achieved by formulating $\Pr(G | D)$  in a novel way 
that relaxes the independence assumption between the blankets.
Essentially, the whole posterior is obtained by computing the posterior of the blanket of each variable
as a conditional distribution that takes into account information from other blankets in the network. 
In the experiments we show that the proposed approximation 
can improve the sample complexity of state-of-the-art scores 
when learning networks with complex topologies, that commonly appear in real-world problems.

After providing some preliminaries, notations and definitions in Section~\ref{sec:stateoftheart},
we introduce the BJP scoring function in Section~\ref{sec:bjp}.
% together with an efficient optimization method to use it.
Section~\ref{sec:experiments} presents the experimental results for several study cases.
Finally, Section~\ref{sec:conclusions} summarizes this work, 
and poses several possible directions of future work.

\section{Background \label{sec:stateoftheart}}
We begin by introducing the notation used for MNs.
Then we provide some additional background about these models
and the problem of learning their independence structure,
and also discuss the state-of-the-art of MN structure learning.
\subsection{Markov networks}
% Let us introduce some notation. 
Have $V$ as a finite set of indexes,
lowercase subscripts for denoting particular indexes, e.g., \(i, j\in V\), 
and uppercase subscripts for subsets of indexes, e.g., \(W\subseteq V\). 
Let $X_V$ be the set of random variables of a domain, denoting single variables as single indexes in $V$, 
e.g.,  $X_i, X_j\in X_V$ when $i,j\in V$. 
% Also, for simplicity, $X$ is used instead of $X_V$. 
% Moreover, the symbol $V$ is overloaded to also denote the set of nodes of an undirected graph $G$. 
% 
% 
For a MN representing a probability distribution $P(X_V)$,
its two components are denoted as follows: $G$, and $\theta$.
% $G$ is a qualitative component that represents
% structural information about a problem domain in the form of independence relationships
% between variables. 
$G$ is the structure, an undirected graph
$G = (V, E)$ where the nodes $V = \{0, . . . , n-1\}$ are the indices of each random variable $X_i$ of the domain,
and $E \subseteq \{V \times V \}$ is the edge set of the graph.
A node $i$ is a neighbor of $j$ when the pair $(i, j) \in E$.
The edges encode direct probabilistic influence between the variables. 
Similarly, the absence of an edge manifests that the dependence could be mediated by some other subset of variables, 
corresponding to conditional independences between these variables. 

A variable $X_i$ is conditionally independent of another non-adjacent variable $X_j$
given a set of variables $X_Z$ if $\Pr(X_i \mid X_j,X_Z) = \Pr(X_i \mid X_Z)$. 
This is denoted by $\ci{X_i}{X_j}{X_Z}$ 
(or $\cd{X_i}{X_j}{X_{Z}}$ for the dependence assertion).
As proven by \cite{hammersley1971markov}, the independences encoded by $G$ allow the decomposition of
the joint distribution into simpler lower-dimensional functions called factors, or potential functions. 
The distribution can be factorized as the 
product of the potential functions $\phi_c(V_c)$
over each clique $V_c$ (i.e., each completely connected sub-graph) of $G$, that is
\begin{equation} \label{eq:Gibbs}
 P(V) = \frac{1}{Z}\prod_{c \in \text{cliques}(G)} \phi_c(V_c),
\end{equation}
where $Z$ is a constant that normalizes the product of potentials.
Such potential functions are parameterized by the set
of numerical parameters $\theta$.

For each variable $X_i$ of a MN, its Markov blanket 
is composed by the set of all its neighbor nodes in the graph.
Hereon we denote the blanket of a variable $X_i$ as $\blanket{X_i}$. 
An important concept that is satisfied by MNs is the Local Markov property, formally described as:
\begin{itemize}
 \item[] \textbf{Local Markov property}. A variable is conditionally independent of all its non-neighbor variables given its MB. 
 That is
 \begin{equation} \label{eq:lmp} %\large
\ci{X_i}{\{X_V \setminus \blanket{X_{i}} \}}{\blanket{X_{i}}}.
 \end{equation}
\end{itemize} 
By using such property, the conditional independences of $P(X_V)$
can be read from the structure $G$. 
This is done by considering the concept of separability. Each pair of non-adjacent variables $(X_i,X_j)$ 
is said to be separated by a set of variables $X_Z \subseteq X_V \setminus \{X_i,X_j\}$
when every path between $X_i$ and $X_j$ in $G$ contains some node in $X_Z$ \cite{pearl88}. 

% The structure in a MN can be characterized by the Markov properties. Here we reproduce
% the Local Markov property, since it is a key concept in this work:
% % \begin{small}
% \begin{enumerate}
% %  \item \emph{Pairwise Markov property}. Any two non-adjacent variables 
% %  are conditional independent given all other variables:
% %  \begin{equation} \label{eq:pmp}% \large
% %  \ci{X_i}{X_j}{V \setminus \{X_i,X_j\}} ~~~\text{if } \{i,j\} \notin E(G).
% %  \end{equation}
%  \item \emph{Local Markov property}. A variable is conditionally independent of all its non-neighbor variables given its Markov blanket
%  \begin{equation} \label{eq:lmp} %\large
% \ci{X_i}{\{V \setminus \blanket{i} \}}{\blanket{X_{i}}}.
%  \end{equation}
% \end{enumerate} 
% \end{small}

In machine learning, statistical independence tests are a well-known tool
to decide whether a conditional independence is supported by the data.
Examples of independence tests used in practice are Mutual Information
\cite{covertomas91}, Pearson's $\chi^2$ and $G^2$ \cite{AGRESTI02}, 
the Bayesian statistical test of independence \cite{MARGARITIS05}, and 
the Partial Correlation test for continuous Gaussian data \cite{Spirtes00}.
Such tests require the construction of a contingency table of counts
for each complete configuration of the variables involved;
as a result, they would have an exponential cost in the number of variables \cite{COCHRAN54}.
For this reason, the use of the local Markov property 
has a positive effect for learning independence structures, allowing the use of smaller tests.
Accordingly, the BJP score introduced in this work takes advantage of this property
by computing a set of conditional probabilities that are more reliable and less expensive.
% This is achieved by examining the irregularities present in a structure.

% 
% It allows to take advantage of the irregularities present in a structure, to 
% infer related independences.
% % for taking advantage of the irregularities present in a structure, improving its posterior estimation.

\subsection{MN structure learning \label{sec:structurelearningalgs}}
The MN structure is learned from a training dataset $D = \{D_1, ..., D_d \}$, 
assumed to be a representative sample of the underlying distribution $P(X_V)$.
Commonly, $D$ has a tabular format, with a column for each variable of the domain $X_V$, 
and one row per data point. 
This work assumes that each variable is discrete, with a finite number of
possible values, and that no data point in $D$ has missing values.
As mentioned in the introduction, this work focuses on methods for computing 
$\Pr(G | D)$. For this reason, 
in this subsection we discuss two recently proposed scoring functions
that approximate it: the Marginal Pseudo Likelihood (MPL) score \cite{pensar2016marginal}, and 
the Independence-based score (IB-score) \cite{schluter2014ibmap}. 

In MPL, each graph is scored 
by using an efficient approximation to the posterior probability of structures given the data.
This score approximates the posterior by considering $P(G \mid D ) \propto P(D \mid G ) \times P(G)$. 
% \end{equation}
Since the data likelihood of the graph, $P(D \mid G)$, is in general extremely hard to evaluate, 
MPL utilizes the well-known approximation called the pseudo-likelihood \cite{besag1972nearest}.
This score was proved to be consistent, that is,
in the limit of infinite data the solution structure has the maximum score.
For finding the MPL-optimal structure, two algorithms were presented:
an exact algorithm using pseudo-boolean optimization, and a fast alternative to the exact method,
which uses greedy hill-climbing with near-optimal performance.
% In our experimentss in Section~\ref{sec:hd} we used the latter, to compare its effi
This algorithm learns the blanket for each variable, locally optimizing 
the MPL for each node, independently of the solutions of the other
nodes. For this, it uses an approximate deterministic hill-climbing procedure 
similar to the well-known IAMB algorithm \cite{tsamardinos03:IAMB}. 
Finally, a global graph discovery method is applied by using a greedy hill-climbing
algorithm, searching for the structure with maximum MPL score, but only restricting the search space 
to the conflicting edges. 
% In our experimental section, such algorithm is referred to as MPL-IAMB-HC.

% \subsubsection{The Independence-based maximum-a-posteriori approach \label{sec:optIBMAP}}
The independence-based score (IB-score) \cite{schluter2014ibmap}
is also based on the computation of the posterior, 
but using the statistics of a set of conditional independence tests.
In this score the posterior $\Pr(G \mid D)$ is computed by combining the outcomes of a set
of conditional independence assertions that completely determine $G$.
Such set was called the \textit{closure} of the structure, denoted $\closure(G)$.
Thus, when using IB-score the problem of structure learning is posed as the maximization of the posterior of 
the closure for each structure. Formally,
\begin{equation} 
 G^{\star} = \arg\max_{G}{\Pr(\closure(G) \mid D)}.  
\end{equation}

Applying the chain rule over the posterior of the closure,
\begin{equation} \label{eq:chainRule} 
    \Pr(\closure(G) \mid D) = \prod_{c_i \in \closure(G)} \Pr( c_i | c_1, \ldots, c_{i-1}, D ),
\end{equation}
the IB-score approximates such probability by assuming 
that all the independence assertions $c_i$ in the closure $\closure(G)$ are mutually independent.
The resulting scoring funtion is computed as:
\begin{equation}\label{eq:independenceApproximationLogs}
 \mbox{IB-score}(G) = \prod\limits_{c_i \in \closure(G)} \log  \Pr( c_i \mid D )  , 
\end{equation}
where each term $\log \Pr(c_i \mid D )$ 
is computed by using the Bayesian statistical test of conditional independence \cite{MARGARITIS05,margaritisBromberg09}.
% Appendix A presents a summary of the formulas used by this statistical test,
% which is also used in the BJP scoring function, proposed in the next section.
% 
Together with the IB-score, an efficient algorithm called IBMAP-HC is presented 
to learn the structure by using a heuristic local search 
over the space of possible structures. 
% IBMAP-HC has proven to improve significantly the quality
% of independence-based competitors.

% Such algorithm utilizes a specific closure set
% to determine each independence structure with a polynomial number of statistical tests.
% In IBMAP-HC the closure set is defined as 
% the union of a set $\closure_X(G)$ of independence and dependence assertions 
% for each variable $X_i$ in the domain $V$, i.e., 
%    \begin{equation} \label{eq:closureMB}
%    \closure(G) = \bigcup\limits_{X_i \in V} \closure_{X_i}(G),
%    \end{equation}
% where each $\closure_{X_i}(G)$ is the union of two mutually exclusive sets of assertions:
%    \begin{eqnarray} \label{eq:closureMBX}
%    \closure_{X_i}(G) = &\Big\{& \cd{X_i}{X_j}{\blanket{i} \!\setminus\! \{X_j\}}  ~:~  X_j \!\in\! \blanket{i}  \Big\} ~ \cup \nonumber \\
%    &\Big\{& \ci{X_i}{X_j}{\blanket{i} }  ~:~ X_j \!\notin\! \blanket{i}  \Big\}.
%    \end{eqnarray}
% That is, for each neighbor of $X_i$ ($X_j\in \blanket{i}$)
% add a conditional dependence assertion between both variables conditioning on $\blanket{i} \setminus \{X_j\}$;
% and for each non-neighbor of $X_i$ ($X_j\notin \blanket{i}$), add a conditional independence assertion between both variables  
% conditioned on $\blanket{i}$.  
% IBMAP-HC has proven to improve significantly the quality
% of independence-based competitors.
% In this work we offer also a comparison of this algorithm against the MPL-IAMB-HC algorithm.
% which currently does not exist in the literature. 

% 
% THE APPROACH
% 
% 
\section{Blankets Joint Posterior scoring function \label{sec:bjp} } % PLEASE, WE NEED A BEST NAME FOR THIS SCORING FUNCTION.. ! 

% the posterior probability of the independence structure of a MN. 
% In particular, BJP has been designed in order to accurately approximate 
% the posterior of structures for cases where the underlying structure contains irregularities. 
% % For this, BJP exploits the fact that each graph $G$ is
% % uniquely specified by its collection of blankets. 
% % The score is explained in detail in the next subsection.
% % Then, we describe how to efficiently learn structures with BJP in Section~\ref{sec:opt}.
% The correctness of BJP is discussed in the Appendix~\ref{app:correctness}.

% \subsection{Blankets Joint Posterior scoring function} 
We introduce now our main contribution,
the Blankets Joint Posterior (BJP) scoring function.
Consider some graph $G$ representing the independence structure of a positive MN.
It is a well-known fact that, by exploiting the graphical properties of such models,
the independence structure can be decomposed as the unique 
collection of the blankets of the variables \cite[Theorem 4.6 on p.~121]{koller09}.
Thus, the computation of the posterior probability of $G$ given a dataset $D$ 
is equivalent to the joint posterior of the collection of blankets of $G$, that is, 
\begin{equation} \label{eq:gDecomposed} \small
\begin{split} 
  \Pr( G \mid D) = \Pr(\blanket{X_0}, \blanket{X_1}, \ldots, \blanket{X_{n-1}} \mid D).
\end{split}
\end{equation}
In contrast with previous works, where the blanket posteriors 
are simply assumed to be independent \cite{pensar2016marginal,schluter2014ibmap}, 
we applied the chain rule to (\ref{eq:gDecomposed}), obtaining
\begin{equation} \small \label{eq:jointBlankets2_withoutOrder}
\begin{aligned}
\Pr (\blanket{X_{0}}, \ldots, \blanket{X_{n-1}} \mid D)&=\mathlarger{\prod\limits_{i=0}^{n-1}}\Pr \left( \blanket{X_{i}} \Bigg| \left\{ \blanket{X_j} \right\}_{j=0}^{i-1} , D \right).
% &\quad \quad \quad \quad \quad \quad =\mathlarger{\prod\limits_{i=0}^{n-1}}\Pr \left( \blanket{X_{i}} \Bigg| \left\{ \blanket{X_j} \right\}_{j=0}^{i-1} , D \right).
\end{aligned}
\end{equation}
In this way, the posterior probability of each blanket
can be described in terms of conditional probabilities, using 
the training dataset $D$ as evidence, together with the blanket of the other variables.
Thus, the joint posterior of all the blankets is computed taking advantage 
of how the blankets are mutually related, instead of assuming them to be independent.
The correctness of the proposed method is discussed in Appendix~\ref{app:correctness}.  
Details about how the BJP scoring function proceeds are presented below.

The computation of $\Pr (\blanket{X_{0}}, \ldots, \blanket{X_{n-1}} \mid D)$ 
has to be done progressively, 
first calculating the posterior of the blanket of a variable,
and then, the knowledge obtained so far can be used 
as evidence to compute the posterior of the blanket of other variables.
However, this decomposition is not unique, since each possible ordering for the variables
is associated to a particular decomposition. 
% By way of example, let us assume that $V=\{X_0,X_1,X_2\}$.
% The six possible decompositions of $\Pr (\blanket{X_{0}}, \blanket{X_{1}}, \blanket{X_{2}} \mid D)$  are:
% 
% \begin{small}
% \begin{eqnarray*}
% \begin{aligned} 
% &\Pr(\blanket{X_{0}} \mid \blanket{X_1}, \blanket{X_2}, D)\times\Pr(\blanket{X_{1}} \mid \blanket{X_2}, D)\times\Pr(\blanket{X_{2}} \mid D) \\
% &\Pr(\blanket{X_{0}} \mid \blanket{X_2}, \blanket{X_1}, D)\times\Pr(\blanket{X_{2}} \mid \blanket{X_1}, D)\times\Pr(\blanket{X_{1}} \mid D) \\
% &\Pr(\blanket{X_{1}} \mid \blanket{X_0}, \blanket{X_2}, D)\times\Pr(\blanket{X_{0}} \mid \blanket{X_2}, D)\times\Pr(\blanket{X_{2}} \mid D) \\
% &\Pr(\blanket{X_{1}} \mid \blanket{X_2}, \blanket{X_0}, D)\times\Pr(\blanket{X_{2}} \mid \blanket{X_0}, D)\times\Pr(\blanket{X_{0}} \mid D) \\
% &\Pr(\blanket{X_{2}} \mid \blanket{X_1}, \blanket{X_0}, D)\times\Pr(\blanket{X_{1}} \mid \blanket{X_1}, D)\times\Pr(\blanket{X_{0}} \mid D) \\
% &\Pr(\blanket{X_{2}} \mid \blanket{X_0}, \blanket{X_1}, D)\times\Pr(\blanket{X_{0}} \mid \blanket{X_1}, D)\times\Pr(\blanket{X_{1}} \mid D). \\
% \end{aligned}
% \end{eqnarray*}
% \end{small}
% 
% 
The basic idea underlying the computation of BJP is to sort the blankets by their size 
(that is, the degree of the nodes in the graph) in ascending order. 
This allows a series of inference steps, in order to avoid the computation of expensive and unreliable probabilities,
thus improving data efficiency.
This is due to the fact that as the size of the blanket increases, greater amounts of data
are required for accurately estimating its posterior probability. 
By using the proposed strategy, the blanket posteriors of variables with fewer neighbors are computed first,
and this information is used as evidence when computing the posteriors for variables with bigger blankets.
As a result, the information obtained from the more reliable blanket posteriors is used
for computing less reliable blankets posteriors.

\begin{figure}[!t]
\centering
\includegraphics[width=.2\textwidth]{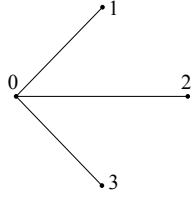}  
\caption{Example of an undirected graph with 4 nodes and hub topology \label{fig:hub}} 
\end{figure}

Now consider an example probability distribution $\Pr(X_V)$ with four variables
$X=\{X_0,X_1,X_2,X_3\}$, represented by a MN whose independence structure $G$ is given 
by the graph of Figure~\ref{fig:hub}. 
% When sorting its nodes by their degree in ascending order, 
% the vector $(X_1,X_2,X_3,X_0)$ can be obtained, and the blankets joint posterior is decomposed as 
One possible way of sorting its nodes by their degree in ascending order is represented by the vector
$(X_1,X_2,X_3,X_0)$, 
and according to this ordering the blankets joint posterior is decomposed as 
\begin{small}
\begin{eqnarray*}
\begin{aligned} \small
\Pr ( \blanket{X_0}, \blanket{X_1}, \ldots, \blanket{X_{n-1}} | D)&= \Pr ( \blanket{X_1} | D ) \\
& \times \Pr ( \blanket{X_2} | \blanket{X_1}, D ) \\
& \times \Pr ( \blanket{X_3} | \blanket{X_1}, \blanket{X_2}, D ) \\
& \times \Pr ( \blanket{X_0} | \blanket{X_1}, \blanket{X_2}, \blanket{X_3},D ).
\end{aligned}
\end{eqnarray*}
\end{small}
% \begin{equation*} 
% \Pr \left( \blanket{X_1} \Bigg| D \right) \times \\
% \Pr \left( \blanket{X_2} \Bigg| \blanket{X_1}, D \right)\times \\
% \Pr \left( \blanket{X_3} \Bigg| \blanket{X_1}, \blanket{X_2}, D \right) \times \\
% \Pr \left( \blanket{X_0} \Bigg| \blanket{X_1}, \blanket{X_2}, \blanket{X_3},D \right).
% \end{equation*}
This example allows us to illustrate the intuition behind BJP, since
the sample complexity of the blanket posterior for variables $X_1$, $X_2$, and $X_3$ is
lower than that of $X_0$. 
Moreover, in this example it is clear that the posterior distribution of $\blanket{X_0}$ is 
not independent of the posterior distributions of $\blanket{X_1}$, $\blanket{X_2}$ and $\blanket{X_3}$.
Clearly, the posterior of $\blanket{X_0}$ is harder to evaluate 
than the posterior of the remaining variables,
and then, computing $\Pr ( \blanket{X_0} | \blanket{X_1}, \blanket{X_2}, \blanket{X_3},D )$ could be more informative
that only computing $\Pr ( \blanket{X_0} | D )$ independently of the rest of blankets.  

Given an undirected graph $G$, denote $\psi$ the ordering vector which contains the variables sorted by their degree in ascending order. 
Therefore, we reformulate \eq{eq:jointBlankets2_withoutOrder} as
\begin{small}
\begin{equation} \label{eq:jointBlankets2}
\begin{aligned} 
BJP(G)&=\mathlarger{\mathlarger{\prod}}\limits_{i=0}^{n-1} 
	  \Pr \left( \blanket{\psi_{i}} \Bigg| \left\{ \blanket{\psi_j} \right\}_{j=0}^{i-1} , D \right).
\end{aligned}
\end{equation}
\end{small}
We now proceed to express the posterior of a blanket in terms of probabilities of conditional independence and dependence assertions. 
The computation of $\Pr ( \blanket{\psi_{i}} | \{ \blanket{\psi_j} \}_{j=0}^{i-1} , D )$
can be derived from the posterior of the independences and dependences represented by each blanket:
\begin{small}
\begin{equation} \label{eq:blanketposterior}
\begin{aligned} 
\Pr \left( \blanket{\psi_{i}} \Bigg| \left\{ \blanket{\psi_j} \right\}_{j=0}^{i-1} , D \right) = 
&~~\mathlarger{\mathlarger{\prod}}\limits_{\psi_k \notin \blanket{\psi_{i}}} \Pr \left( \ci{\psi_{i}}{\psi_{k}}{\blanket{\psi_{i}}} \Bigg| \left\{ \blanket{\psi_j} \right\}_{j=0}^{i-1} , D \right) \times \\
&~~\mathlarger{\mathlarger{\prod}}\limits_{\psi_k \in \blanket{\psi_{i}}} \Pr \left( \cd{\psi_{i}}{\psi_{k}}{\blanket{\psi_{i}} \setminus \{ \psi_{k}\} } \Bigg| \left\{ \blanket{\psi_j} \right\}_{j=0}^{i-1} , D \right).
\end{aligned}
\end{equation}
\end{small}
In this way, the whole score is the product of the posterior probability 
of each blanket, computed in terms of posterior probabilities conditioned in other blankets.
The particular way of determining the posterior of each blanket of (\ref{eq:blanketposterior})
is inspired by the \textit{Markov blanket closure} \cite[Definition~2]{schluter2014ibmap}, which 
is a set of independence and dependence assertions 
formally proven to determine a MN structure.

The two factors in  (\ref{eq:blanketposterior}) will be interpreted as follows: 
\begin{itemize}
 \item The first product computes the probability of independence between $\psi_i$ 
and its non-adjacent variables, conditioned on its blanket,
given the previously computed blankets and the dataset $D$.
It can be computed as
\begin{small}
\begin{equation}  \label{eq:ciPosterior}
\begin{aligned}
\Pr\left( \ci{\psi_{i}}{\psi_{k}}{\blanket{\psi_{i}}}\Bigg|\left\{\blanket{\psi_j}\right\}_{j=0}^{i-1},D \right)
\quad = \begin{cases}
\Pr(\ci{\psi_{i}}{\psi_{k}}{\blanket{\psi_{i}}}\mid D)\\ 
 &\hspace{-1.4cm}\text{if $i < k$,}\\
 ~~\\
 1  & \hspace{-1.4cm}\text{if $i > k$. }
\end{cases}
\end{aligned}
\end{equation}
\end{small}

Here, $i < k$ indexes over the variables for which the blanket posterior probability is not already computed.
For the remaining variables the posterior of independence will be simply inferred as 1. 
With this strategy, the score simply uses the information 
in the evidence, since the independence is determined by the blanket of $\psi_k$.
The rationale behind this inference is that for cases $i>k$, 
the blanket of $\psi_{k}$ has already been computed. 
As it will be proved in more detail in Appendix~\ref{app:correctness}, 
$\blanket{\psi_{k}}$ already contains information about the independence of $\psi_{i}$ and $\psi_{k}$ .
By considering the local Markov property for the blanket of 
$\psi_k$, and the fact that $\psi_k$ is not in the blanket of $\psi_i$, 
the opposite must also be true (as these are undirected edges).
\item The second product in \eq{eq:blanketposterior} computes the posterior probability of dependence 
between $\psi_i$ and its adjacent variables, conditioned on its remaining neighbors,
given the blankets computed previously and the dataset $D$.
It can be computed as 
\begin{small}
\begin{equation}  \label{eq:cdPosterior}
\begin{aligned}
\Pr\left(\cd{\psi_{i}}{\psi_{k}}{\blanket{\psi_{i}}\setminus\{\psi_{k}\}}\Bigg|\left\{\blanket{\psi_j}\right\}_{j=0}^{i-1},D\right)
 = \begin{cases}
\Pr(\cd{\psi_{i}}{\psi_{k}}{\blanket{\psi_{i}}\setminus \{\psi_{k}\}}\mid D)\\ 
 &\hspace{-1.4cm}\text{if $i < k$,}\\
 ~~\\
 1  & \hspace{-1.4cm}\text{if $i > k$. }
\end{cases}
\end{aligned}
\end{equation}
\end{small}
Here, again $i < k$ indexes over the variables for which the blanket posterior is not already computed.
For the remaining variables the posterior of dependence will be inferred as 1. 
Again, the score use the evidence information, since the independence is determined by the blanket of $\psi_k$.
% Also, this inference can be done since the dependence is determined by the blanket of $\psi_k$, 
% which is in the evidence $\left\{ \blanket{\psi_j} \right\}_{j=0}^{i-1}$. 
% The correctness of this inference step is also discussed in Appendix~\ref{app:correctness}.
\end{itemize}
For the sake of clarity, Appendix \ref{appendix:example} shows 
the complete computation of the BJP score for the graph of Figure~\ref{fig:hub}.

The only approximation in BJP is made in (\ref{eq:blanketposterior}), 
by assuming that all the independence and dependence assertions that determine the blanket of a variable $\psi_i$
are mutually independent. This is a common assumption, 
made implicitly by all the constraint-based MN structure learning algorithms
\cite{schluter2012survey}, and also by the MPL score and the IB-score. 
For the computation of the posterior probabilities of independence 
$\Pr( \ci{\psi_{i}}{\psi_{k}}{\blanket{\psi_{i}}} \mid D)$
and dependence $\Pr( \cd{\psi_{i}}{\psi_{k}}{\blanket{\psi_{i}}  \setminus \{ \psi_{k}\}   } \mid D )$
used in \eq{eq:ciPosterior} and \eq{eq:cdPosterior}, respectively,
BJP uses the Bayesian test of \cite{margaritisBromberg09,MARGARITIS05,MARGARITIS00}, 
in the same way as the IB-score explained in the previous section. 
Precisely, this statistical test computes the posterior of independence and dependence assertions, 
and has been proven to be statistically consistent in the limit of infinite data. 

We now discuss the computational complexity of the score. 
For a fixed structure, 
the computational cost is directly determined 
by the number of statistical tests that it is required to perform on data.
Recall that the computational cost of each test is lineal
in the number of variables involved and the number of data points \cite{COCHRAN54}. 
As stated in \eq{eq:jointBlankets2}, BJP 
computes the posterior probability of the blanket for the $n$ variables of the domain.
For each, it is required to perform $n-1$ statistical tests on data, by using \eq{eq:blanketposterior}. 
Then, one half of the tests are inferred when computing the posterior of independences and dependences 
of \eq{eq:ciPosterior}
and \eq{eq:cdPosterior}. 
Thus, only $\frac{n(n-1)}{2}$ tests are required for computing the BJP score of a structure.

% Appendix A presents a summary of these formulas needed to compute such probabilities,
% extracted from \cite{margaritisBromberg09}.

% \subsection{Optimization \label{sec:opt}}

We end this section with the optimization proposed in this work for learning the structure with the BJP score. 
% For learning the structure with the BJP score, 
The na\"ive optimization consists in maximizing over all the possible undirected graphs
for some specific problem domain, as in \eq{eq:maxG},
computing with \eq{eq:jointBlankets2} the score for each structure.
Since the discrete optimization space of the possible graphs $\mathcal{G}$ grows rapidly
with the number of variables $n$, the search is clearly intractable even for small domain sizes.
Hence, in this work we test the performance of BJP 
with brute force only for small domains. For larger domains
we use the IBMAP-HC algorithm, as an efficient approximate solution 
proposed in \cite{schluter2014ibmap}.
% ,in order to get satisfactory quality in a reasonable time. 

The optimization made by IBMAP-HC is a simple heuristic hill-climbing procedure.
The search is initialized by computing the score 
for an empty structure with no edges, and $n$ nodes.
The hill-climbing search starts with a loop that iterates 
by selecting the next candidate structure at each iteration. 
A na\"ive implementation of hill-climbing would select the neighbor structure with maximum score,
computing the score for the 
$n \choose 2$ neighbors that differ in one edge.
Such expensive computation is avoided by selecting the next candidate 
with a heuristic that flips the most promising edge (i.e., the edge with lower local contribution to the score). 
Once the next candidate is selected, its score is computed
to be compared to the best scoring structure found so far. 
The algorithm stops when the neighbor proposed does not improve the current score.
% The computational cost of using this optimization method 
% is determined by the number of ascents done until termination.

% Our experimental results show that using BJP together with 
% this optimization allow us to improve the quality of the structures learned,
% also improving the running time againts the competitor scores. 

% At the end of this work, and as part of the conclusions,
% the results prove that by using BJP the quality of the structures
% can be improved when the problem has irregular structures. 

\section{Experimental evaluation \label{sec:experiments}}	

This section presents several experiments in order to determine the merits of BJP in practical
terms. 
We compare BJP against two recently proposed scoring functions
that approximate the posterior of structures: the Marginal Pseudo Likelihood (MPL) score \cite{pensar2016marginal}, and 
the Independence-based score (IB-score) \cite{schluter2014ibmap}. 
To the best of our knowledge, 
there are no other scoring functions in the literature of MNs for scoring graphical 
independence structures by using $P(G \mid D)$. 

Two sets of experiments are presented, one from low-dimensional problems, 
and another for high-dimensional problems.
For the low-dimensional setting, we used brute force (i.e., exhaustive search) to study the convergence of the
scoring functions to the exact solution. 
The goal is to prove experimentally that the sample complexity
for successfully learning the exact structure of BJP can be better than for the competitors. 
For the high-dimensional setting, we used hill-climbing optimization for all the scoring functions.
This experiments were performed in order to prove that, by using a similar search strategy, 
BJP can identify structures with fewer structural
errors than the competitor scores. 
% Also, a comparison with state-of-the-art structure learning algorithms is presented. 
The software to carry out the experiments has been developed in
Java, and it is publicly available\footnote{http://dharma.frm.utn.edu.ar/papers/bjp}.

For the experiments we selected a set of networks where the topologies exhibit irregularities,
which is a common property in many real-world networks \cite{silva2016machine}.
According to \cite{albertson1997irregularity}, the irregularity of an undirected graph
can be computed by summing the imbalance of its edges: 
  \begin{equation} \label{eq:irregularity}
   irr(G) = \sum_{(i,j) \in E(G)} |d_G(i) - d_G(j)|,
  \end{equation}
where $d_G(i)$ is the degree of the node $i$ in that graph. 
Clearly $irr(G)=0$ if and only if $G$ is regular. 
For non-regular graphs $irr(G)$ is a measure of the lack
of regularity. 
Since BJP can infer complex statistical tests from other more simpler tests performed before, 
we used the irregularity of the underlying structure as an external control variable 
that determines how important is the independence assumption between blankets for decomposable scores. 

% 
% it is a common situation in the practice to have 
% Estas estructuras resultan interesantes porque es muy normal que la irregularidad ocurra en la practica, 
% y porque mientras mas irregulares son las estructuras,
% más drástica se vuelve la suposicion de independencia entre blankets.

% Although there are more complex measures of irregularity 
% for undirected graphs \cite{rautenbach2006extremal,dimitrov2014total},
% this na\"ive definition will suffice for the purposes of this work. 

% Complex network structures describe a wide variety of systems of technological and intellectual importance,
% such as the Internet, World Wide Web, coupled biological and chemical systems, financial, social, neural and communication networks. 

\subsection{Consistency experiments}
A MN scoring function is consistent when the structure which maximizes the score over all the possible structures 
is the correct one, in the limit of infinite data.
However, in practice the data is often too scarce to satisfy this condition,
and the sample size needed to reach the correct structure 
varies across different scoring functions.
This is referred to as the \textit{sample complexity} of the score. 
The experiments here presented were carried out
in order to measure the sample complexity 
of the three different scoring functions known to compute the posterior of structures: MPL, IB-score and BJP.
This is achieved by measuring their ability to return, by brute force, the exact independence structure of the MN
which generated the data.

\captionsetup[subfigure]{labelformat=empty}
\begin{figure}[!htbp]
    \centering
    \begin{subfigure}[b]{0.3\textwidth}
        \includegraphics[width=.8\textwidth]{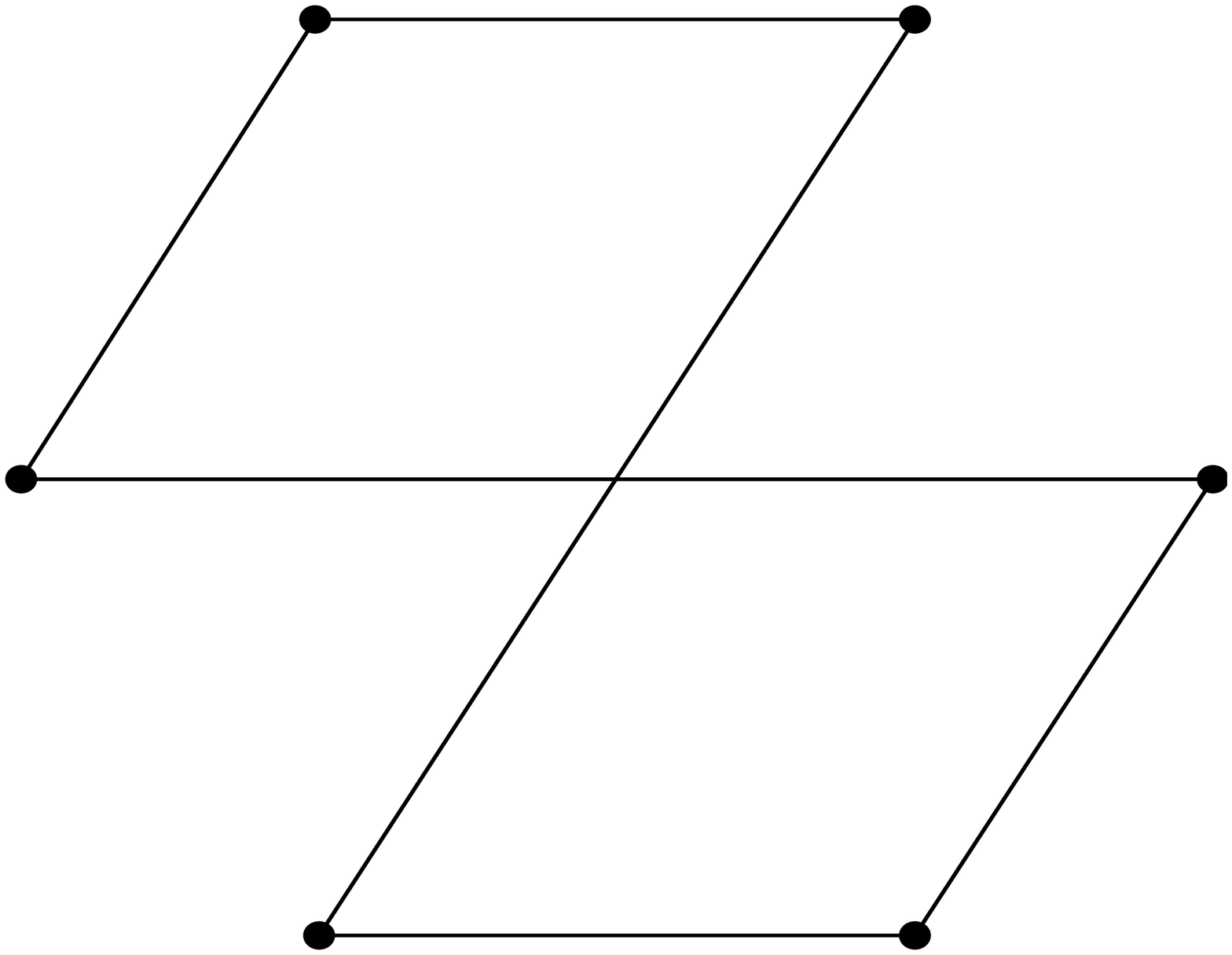}
        \caption{\textbf{Target structure 1} }
        \label{fig:model1}
    \end{subfigure}
    ~ %add desired spacing between images, e. g. ~, \quad, \qquad, \hfill etc. 
    \begin{subfigure}[b]{0.3\textwidth}
        \includegraphics[width=.8\textwidth]{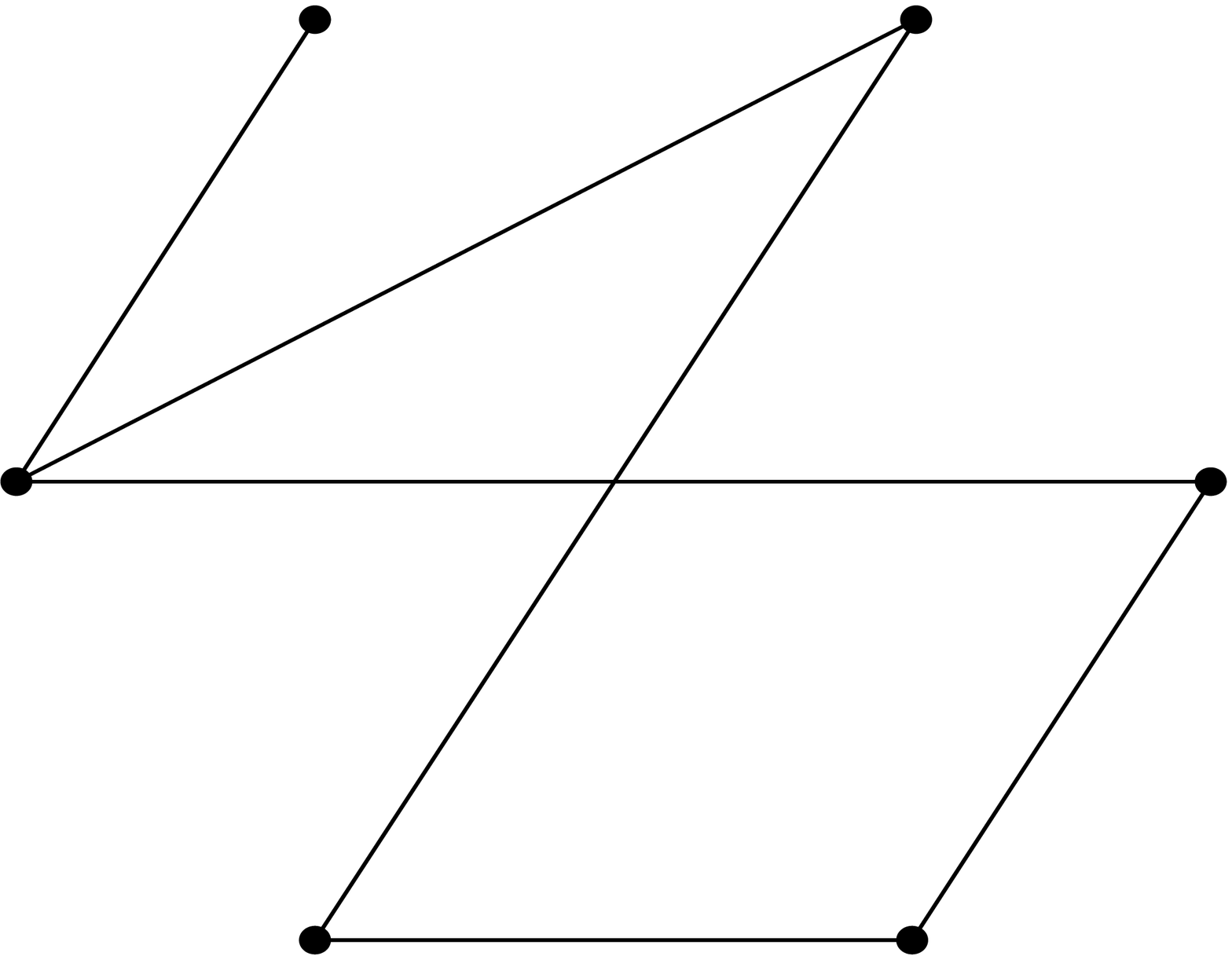}
        \label{fig:model2}
        \caption{\textbf{Target structure 2}}
    \end{subfigure}
    \begin{subfigure}[b]{0.3\textwidth}
        \includegraphics[width=.8\textwidth]{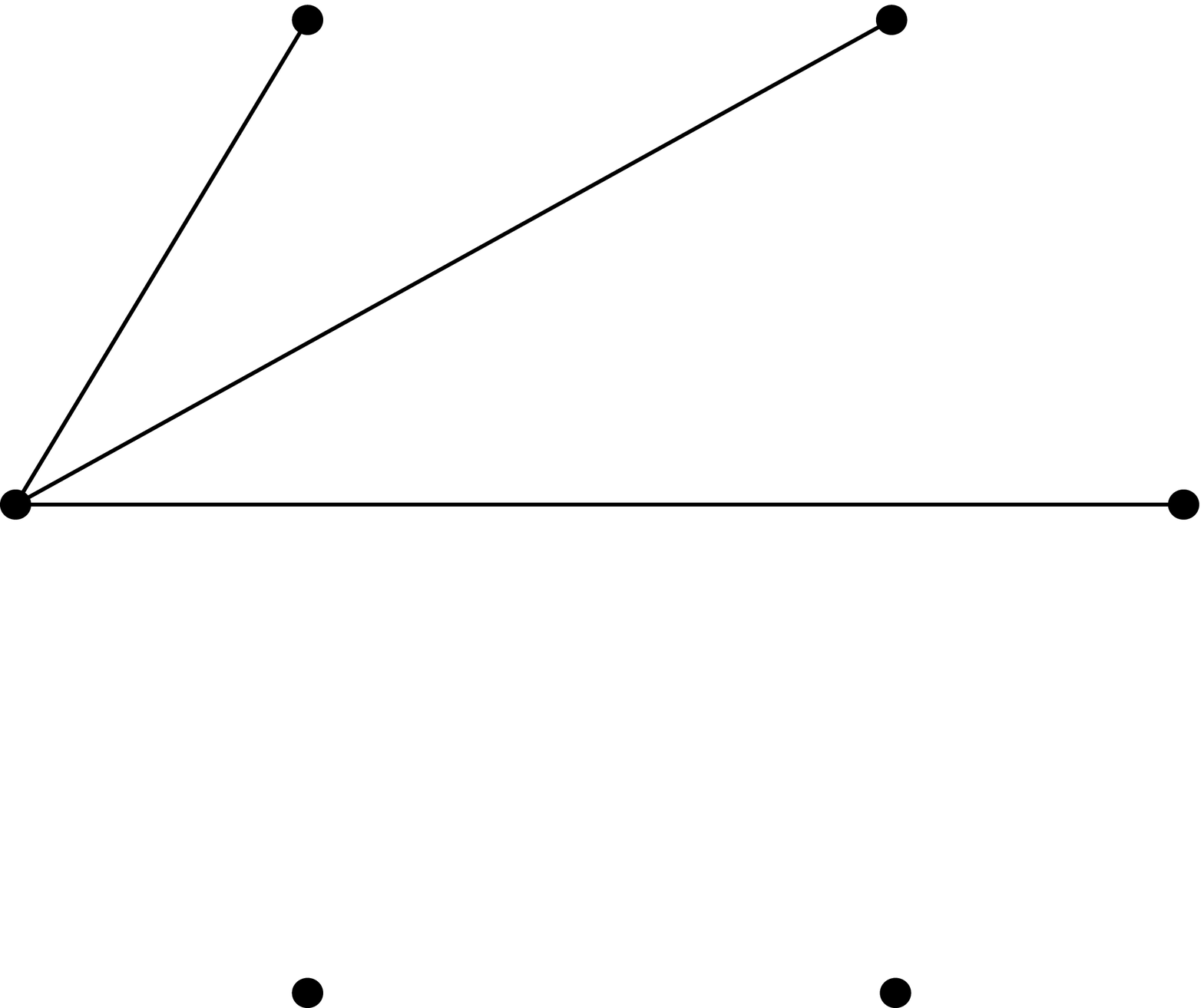}
        \caption{\textbf{Target structure 3}}
        \label{fig:model3}
    \end{subfigure}
\\  
%     ~ \\
    \begin{subfigure}[b]{0.3\textwidth}
        \includegraphics[width=.75\textwidth]{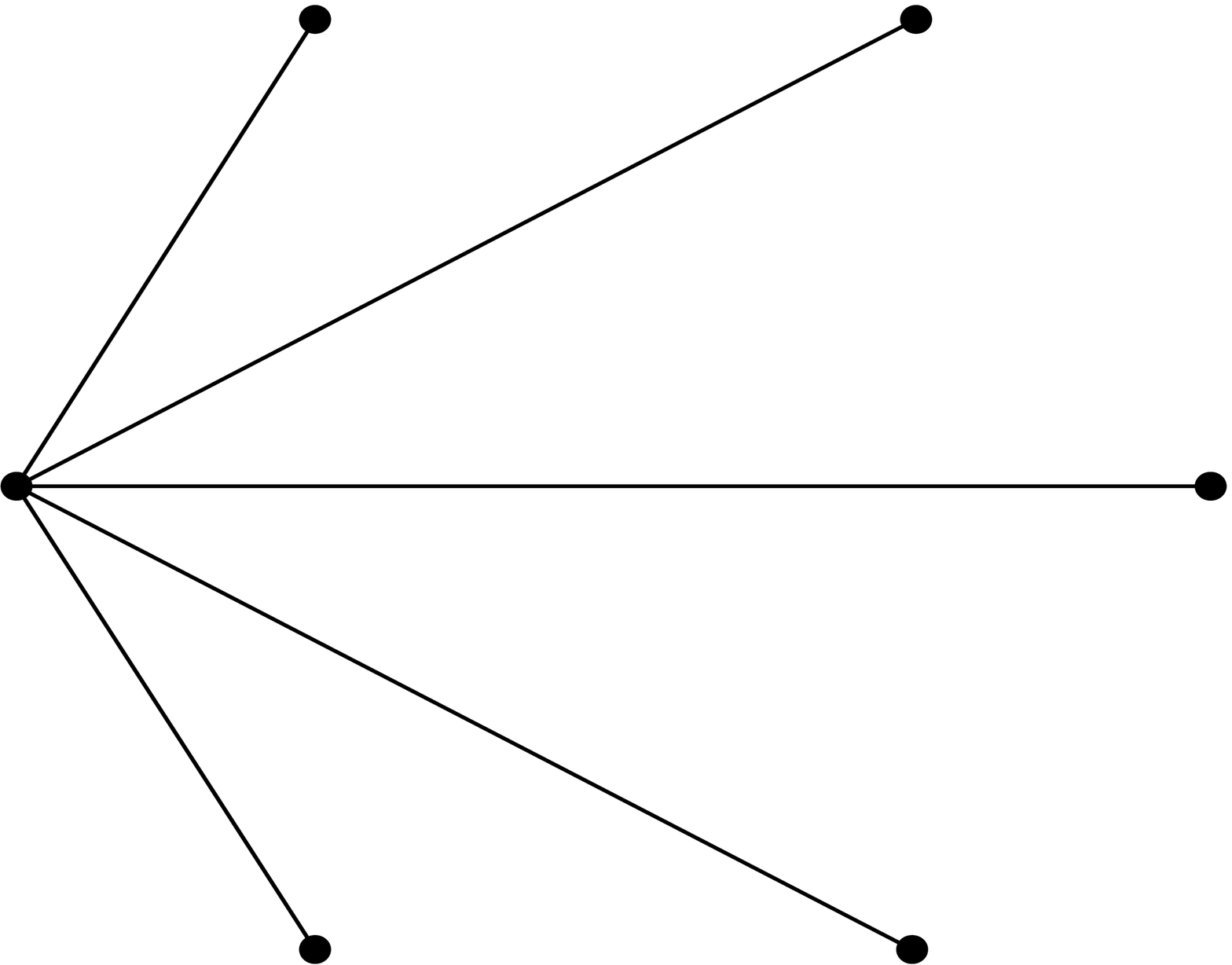}
        \caption{\textbf{Target structure 4}}
        \label{fig:model4}
    \end{subfigure}
    \begin{subfigure}[b]{0.3\textwidth}
        \includegraphics[width=.75\textwidth]{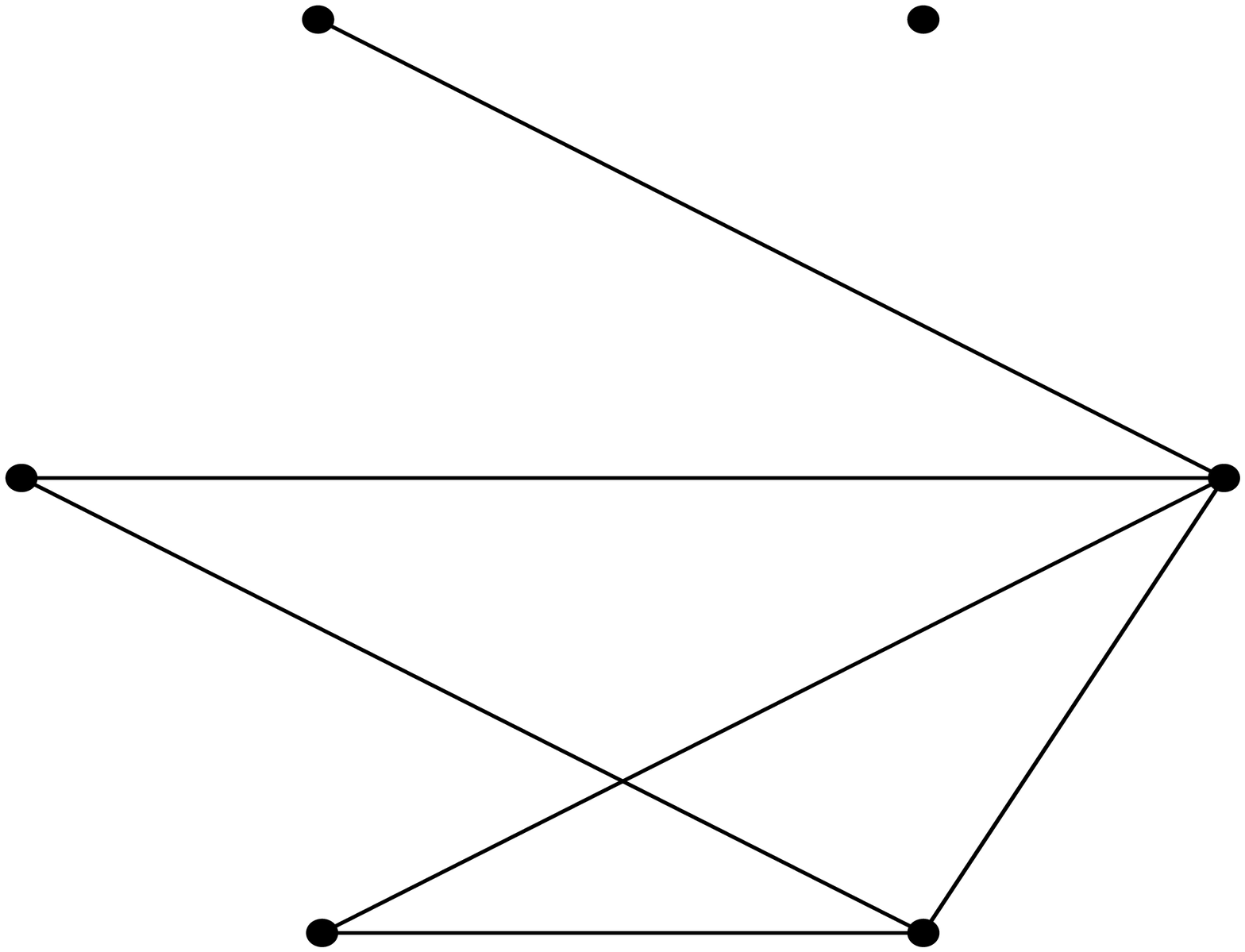}
        \caption{\textbf{Target structure 5}}
        \label{fig:model5}
    \end{subfigure}    
%     ~ %add desired spacing between images, e. g. ~, \quad, \qquad, \hfill etc. 
    \begin{subfigure}[b]{0.3\textwidth}
        \includegraphics[width=.75\textwidth]{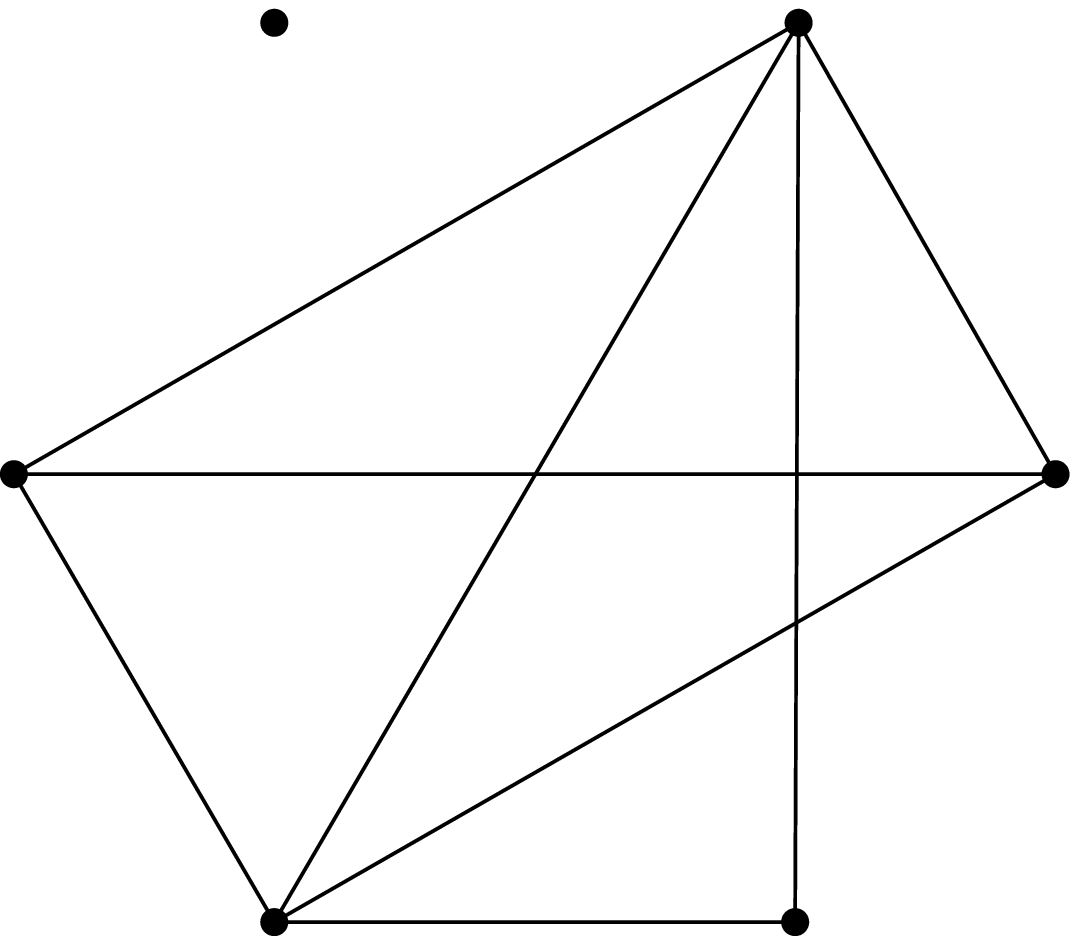}
        \caption{\textbf{Target structure 6}}
        \label{fig:model6}
    \end{subfigure}    
    \caption{Independence structures for the first set of experiments: model 1 is regular ($irr=0$); model 2 has $irr=10$; 
    model 3 has $irr=18$; model 4 has $irr=20$; models 5 and 6 have the maximum irregularity for six variables ($irr= 26$).}\label{fig:n6graphs}
\end{figure}

To make this comparative study, we selected the six different target structures shown in Figure~\ref{fig:n6graphs}. 
These graphs represent different cases of irregularity, according to \eq{eq:irregularity}. 
The first target structure is regular (irr = 0), the second has a little irregularity,
the third and fourth structures are irregular structures with a hub topology, 
and the fifth and sixth target structures have maximum irregularity for $n=6$. 
As mentioned before, the irregularity is used here as a parameter 
for determining how important is the independence assumption between blankets for decomposable scores. 
Thus, in terms of sample complexity, we expect larger improvements of BJP over the competitors 
when the irregularity of the underlying structure increases. 

For constructing a probability distribution from these independence structures according to \eq{eq:Gibbs},
random numeric values were assigned to their maximal clique factors, sampled independently 
from a uniform distribution over $(0,1)$. 
Ten distributions were generated for each target structure, considering only binary discrete variables. 
Then, for each one, ten different random seeds were
used to obtain $100$ datasets for each graph, 
by using the Gibbs sampling tool of the open-source Libra toolkit \cite{lowd2015libra}.
The Gibbs sampler was run with 100 burn-in and 1000 sampling iterations, as commonly used in other works \cite{lowd14a,schluter2014ibmap,van2012markov}. 

Since we have $n=6$ variables, the search space consists of
$2^{6\choose 2}=32768$ different undirected graphs. 
The experiment consisted of evaluating the
number of true structures returned by each score over the 100 datasets. 
This is called here the success rate of the scoring function. 
The success rate is computed 
for increasing dataset sizes $\mathcal{N}_D=\{250,500,1000,2000,4000,8000\}$. 
Of course, since greater sizes of the dataset lead to better estimations,
$\mathcal{N}_D$ affects the quality of the structure learned. 
Therefore, a score is considered better than another score 
when its success rate converges to $1$ with lower values of $\mathcal{N}_D$.
% Also, we report the number of inconsistent structures,
% counted as the number of structures which the scoring function rates higher than the solution
% structure. We consider this measure for comparing the sample complexity
% of different scores, since a value that converges to 0 with a lower amount of data indicates better
% sample complexity. 

% In this experiment, for MPL 
% the equivalent sample size parameter $N = 1$ is set, 
% as used by its authors in a similar experiment. 
% 

\begin{table}[t!]
\scriptsize
\centering
\scalebox{0.9}{
\begin{tabular}{c c c c c c c}
%  \hline
\multicolumn{2}{c}{\textbf{Target}} &\textbf{$Irr$} & \textbf{$\mathcal{N}_D$ }& \multicolumn{3}{c}{\textbf{Success rate}} \\ %\cline{5-7} 
\multicolumn{2}{c}{\textbf{structure}} & & & \textbf{MPL} & \textbf{IB-score } & \textbf{BJP} \\ \hline 

\multirow{6}{*}{\textbf{1}} & \multirow{6}{*}{\includegraphics[width=2cm]{generatingModel3_n6.eps}} 
         & \multirow{6}{*}{0} & 250  & 0.00 & 0.00          & 0.00 \\
			  & & & 500  & 0.00 & 0.00          & \textbf{0.01} \\
			  & & & 1000 & 0.01 & \textbf{0.05} & 0.03 \\
			  & & & 2000 & 0.04 & \textbf{0.15} & 0.12 \\
			  & & & 4000 & 0.15 & \textbf{0.25} & 0.21 \\
			  & & & 8000 & 0.28 & \textbf{0.35} & 0.34  \\ \hline
\multirow{6}{*}{\textbf{2}} & \multirow{6}{*}{\includegraphics[width=2cm]{generatingModel5_n6.eps}} 
        & \multirow{6}{*}{10} & 250  & 0.00 & 0.00          & 0.000  \\
			  & & & 500  & 0.00 & 0.00          & \textbf{ 0.01}\\
			  & & & 1000 & 0.00 & \textbf{0.04} & 0.02 \\
			  & & & 2000 & 0.02 & 0.15          & \textbf{0.16} \\
			  & & & 4000 & 0.10 & \textbf{0.27} & 0.25 \\
			  & & & 8000 & 0.18 & 0.39          & 0.39  \\ \hline
\multirow{6}{*}{\textbf{3}} & \multirow{6}{*}{\includegraphics[width=2cm]{generatingModel9_n6.eps}} 
        & \multirow{6}{*}{18} & 250  & 0.00 & \textbf{0.06} & 0.04  \\
			  & & & 500  & 0.03 & 0.09 & \textbf{0.12}  \\
			  & & & 1000 & 0.10 & 0.17 & \textbf{0.19}  \\
			  & & & 2000 & 0.17 & 0.22 & \textbf{0.27}  \\
			  & & & 4000 & 0.22 & 0.45 & \textbf{0.49} \\
			  & & & 8000 & 0.34 & 0.58 & \textbf{0.61} \\ \hline
\multirow{6}{*}{\textbf{4}} & \multirow{6}{*}{\includegraphics[width=2cm]{generatingModel2_n6.eps}} 
         & \multirow{6}{*}{20}& 250  & 0.00 & 0.00          & 0.00 \\
			  & & & 500  & 0.00 & \textbf{0.03} & 0.02 \\
			  & & & 1000 & 0.00 & 0.06           & \textbf{0.10}  \\
			  & & & 2000 & 0.00 & 0.14           & \textbf{0.18} \\
			  & & & 4000 & 0.00 & 0.29           & \textbf{0.36} \\
			  & & & 8000 & 0.00 & 0.44           & \textbf{0.50}  \\ \hline
\multirow{6}{*}{\textbf{5}} & \multirow{6}{*}{\includegraphics[width=2cm]{generatingModel1_n6.eps}}
        & \multirow{6}{*}{26} & 250  & 0.00 & 0.01          & 0.01 \\
			  & & & 500  & 0.00 & \textbf{0.02}  & 0.01  \\
			  & & & 1000 & 0.00 & 0.10           & \textbf{0.11} \\
			  & & & 2000 & 0.00 & 0.23           & \textbf{0.26} \\
			  & & & 4000 & 0.03 & \textbf{0.56}  & 0.54 \\
			  & & & 8000 & 0.21 & 0.75           & \textbf{0.76} \\ \hline
\multirow{6}{*}{\textbf{6}} & \multirow{6}{*}{\includegraphics[width=2cm]{generatingModel7_n6.eps}} 
         & \multirow{6}{*}{26}& 250 & 0.00 & 0.00           & 0.00            \\
			  & & & 500 & 0.00 & 0.00           & 0.00            \\
			  & & & 1000 & 0.00 & 0.04          & \textbf{0.13}   \\
			  & & & 2000 & 0.00 & 0.28          & \textbf{0.37}   \\
			  & & & 4000 & 0.02 & \textbf{0.66} & 0.61            \\
			  & & & 8000 & 0.27 & 0.80 & \textbf{0.82}            \\ %\hline
\end{tabular} }
\caption{Success rate of BJP, IB-score and MPL over 100 datasets for the target structures on Figure~\ref{fig:n6graphs}. 
Rates in bold face correspond to the best case. \label{table:consistency}}
\end{table} 

Table~\ref{table:consistency} shows the results of the experiment. 
The first column shows the target structures, the second shows their irregularity,
the third shows each sample size $\mathcal{N}_D$ used,
and the fourth shows the success rate.
For all the cases, it can be seen how the success rate of the three scoring functions grows 
with the sample size $\mathcal{N}_D$.
The results in the fourth column show that BJP 
has a better success rate in almost all cases. %for irregular structures.
For all the cases, MPL has a slower convergence than IB-score and BJP.
This is interesting, since MPL has not been compared before with other approximations of $\Pr(G | D)$,
and the experimental results shown in \cite{pensar2016marginal} 
only compares the quality obtained by using the score with a local hill-climbing search mechanism 
against standard constraint-based algorithms. 
For structures 1 and 2, IB-score shows better convergence than BJP,
but they would eventually converge similarly for greater $\mathcal{N}_D$ sizes.
This is an expected result, because these structures are regular, 
and the approximation of BJP and IB-score are very similar for computing $\Pr(G | D)$.
In contrast, for structures 3, 4, 5 and 6, BJP has in general the best success rate.
This is also an expected result, according to the irregularity of  the underlying structures.
Accordingly, the best improvement of BJP over IB-score is for model 6 
(which has maximal irregularity) and $\mathcal{N}_D = \{1000,2000\}$, 
with an improvement of success rate of up to 9\%.
When compared with MPL, BJP obtains the best improvement in success rate of up to 59\%, also for model 6 and 
$\mathcal{N}_D = \{4000\}$. 

In general, these results are consistent with the hypothesis of this work,
since BJP has been designed to improve the computation of $\Pr(G|D)$, and the irregularity
highlights the cases where an improvement of the sample complexity is expected, due to the independence assumption between blankets 
made by the state-of-the-art scores. The following section shows the performance of the three scoring functions
for more complex domains.

\subsection{Structural errors analysis \label{sec:hd}}
In this section, experiments in the higher-dimensional setting are presented.
For this, we evaluate the quality of the structures learned by using an approximate search mechanism.
The BJP score and the IB-score were tested with the IBMAP-HC algorithm proposed in \cite{schluter2014ibmap}, 
briefly explained at the end of Section~\ref{sec:bjp}.
The MPL scoring function was tested with the most efficient optimization algorithm proposed 
in \cite{pensar2016marginal}, described in Section~\ref{sec:structurelearningalgs}.

\begin{figure}[t!]
\centering
\scriptsize
\textbf{Hub 1}
\includegraphics[width=.166\textwidth]{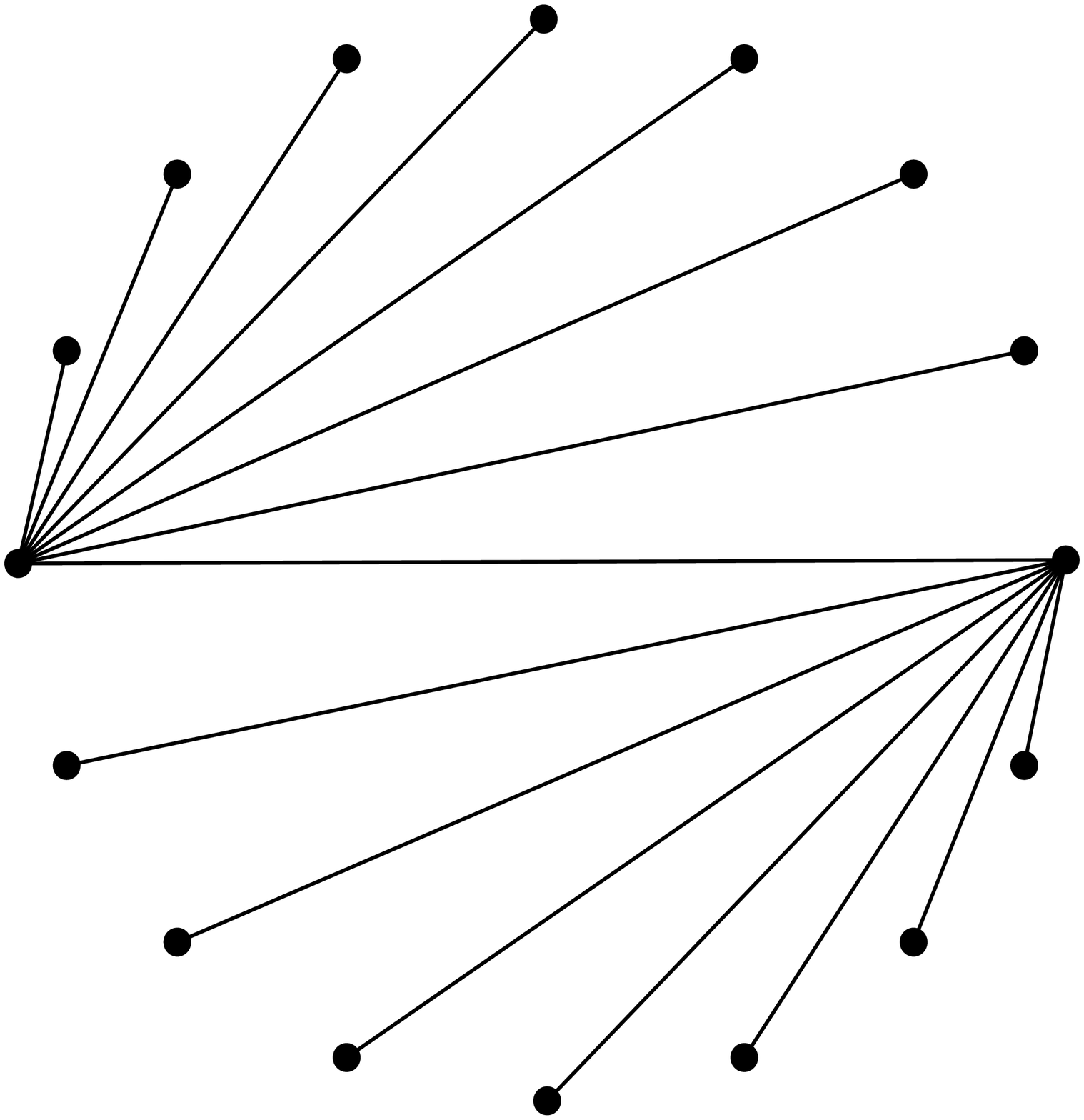}
\textbf{Hub 2}
\includegraphics[width=.166\textwidth]{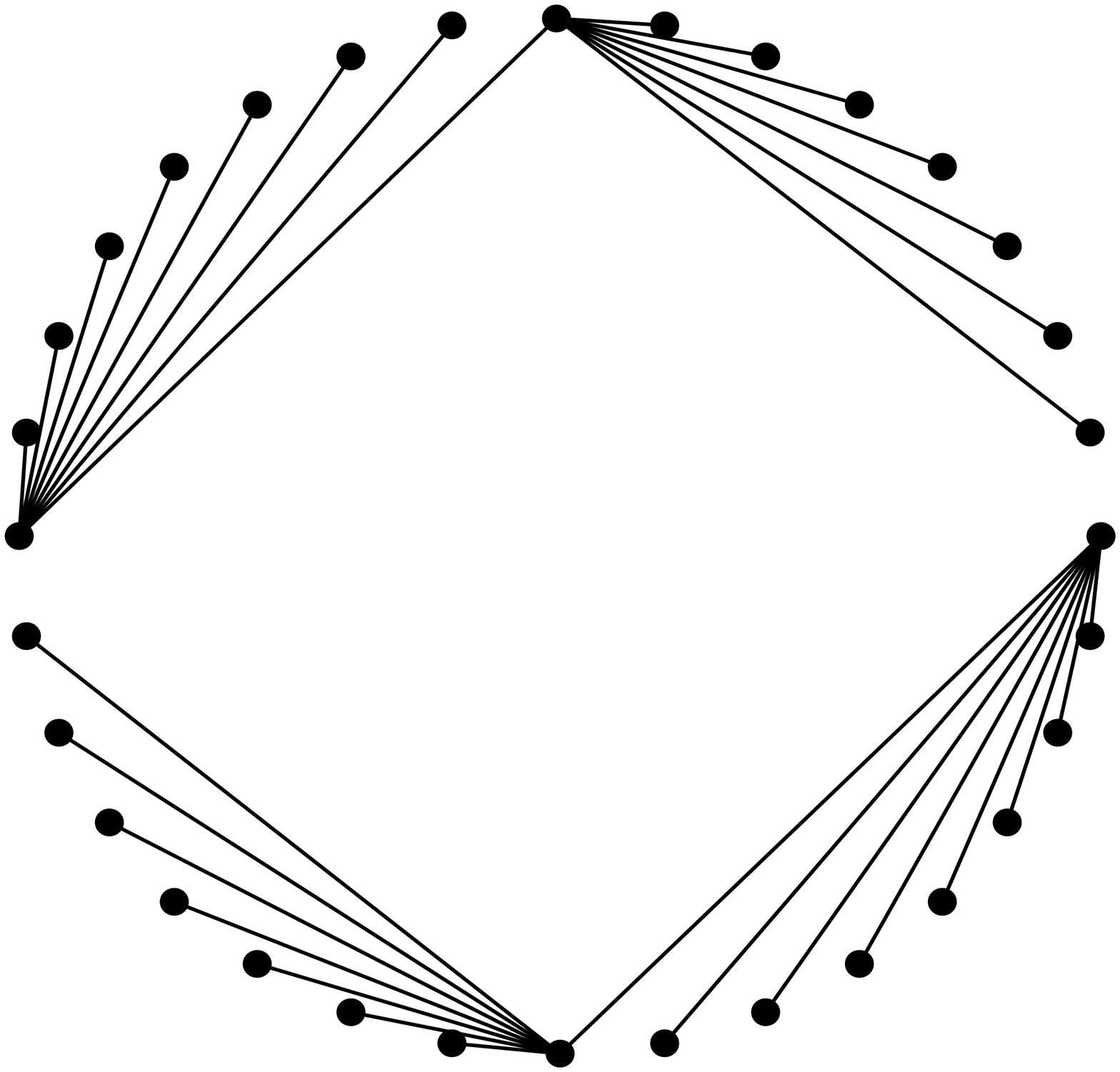}  
\textbf{Hub 3}
\includegraphics[width=.166\textwidth]{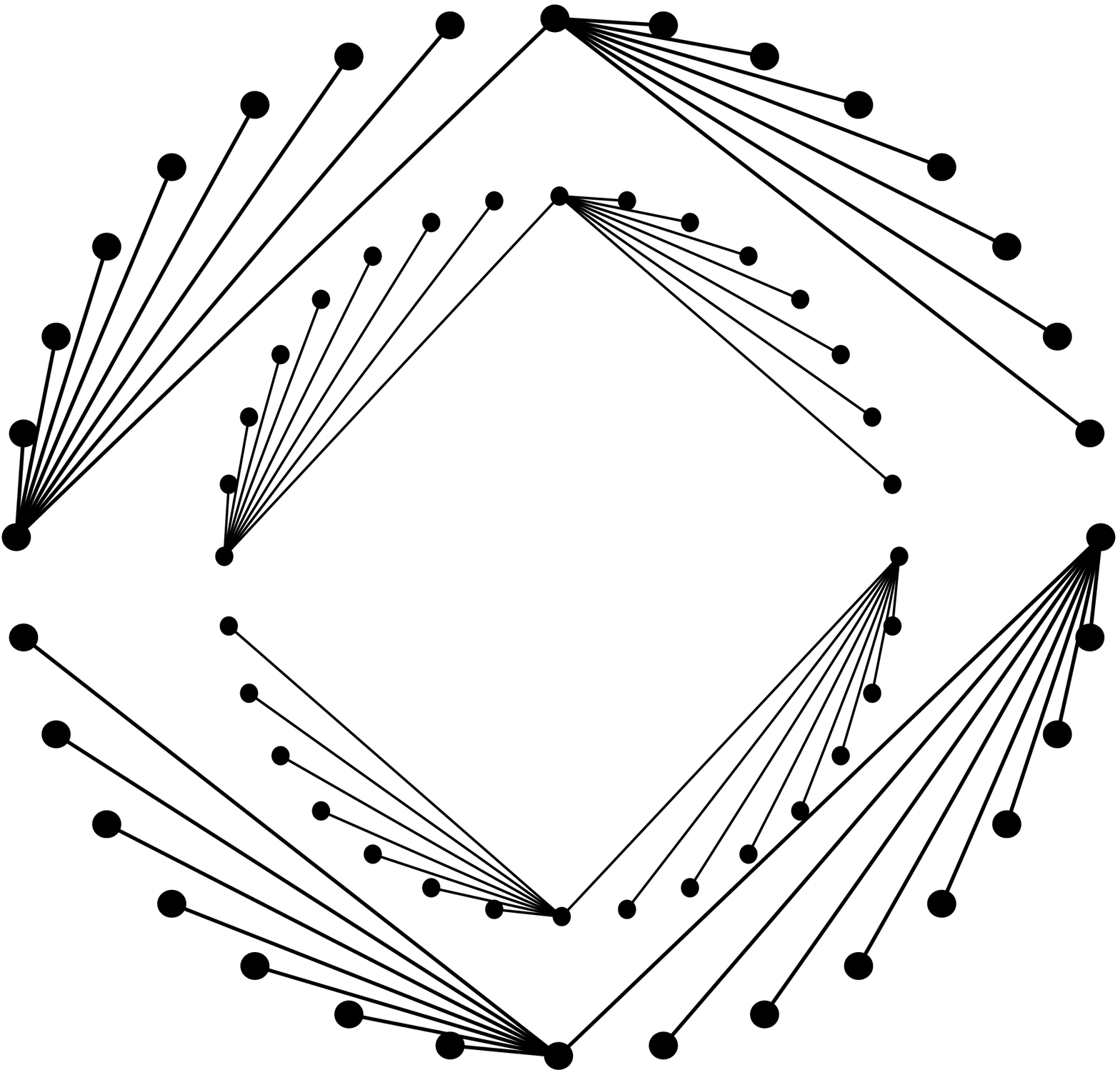}  
\textbf{Hub 4}
\includegraphics[width=.166\textwidth]{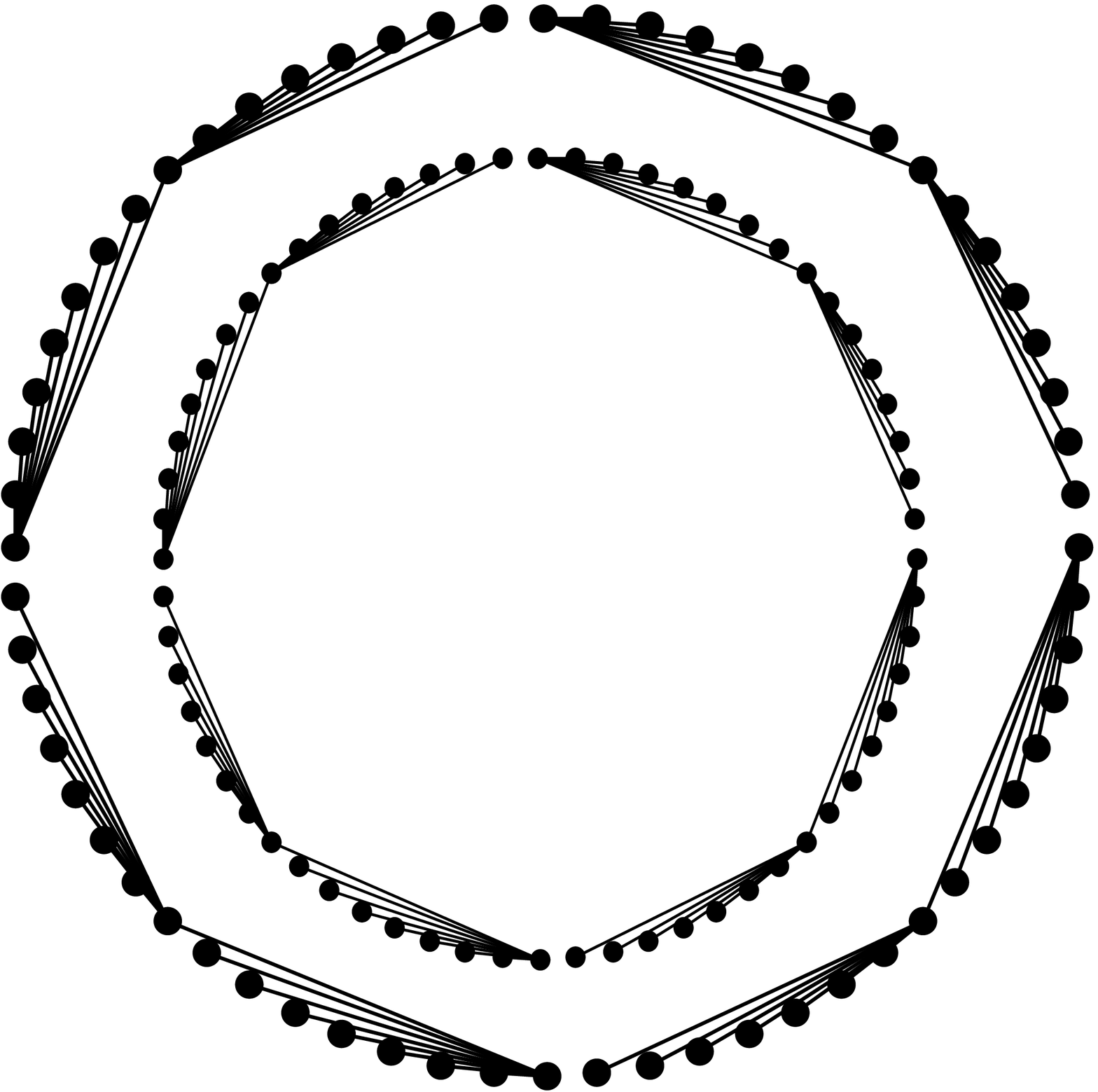} \\ ~~ \\  
\caption{Structures with a hub topology and 16, 32, 64 and 128 nodes \label{fig:hubs}} 
\end{figure}

\begin{figure}[b!]
\scriptsize
% \centering
\textbf{Scale-free 1}                                         
\includegraphics[height=.2\textwidth]{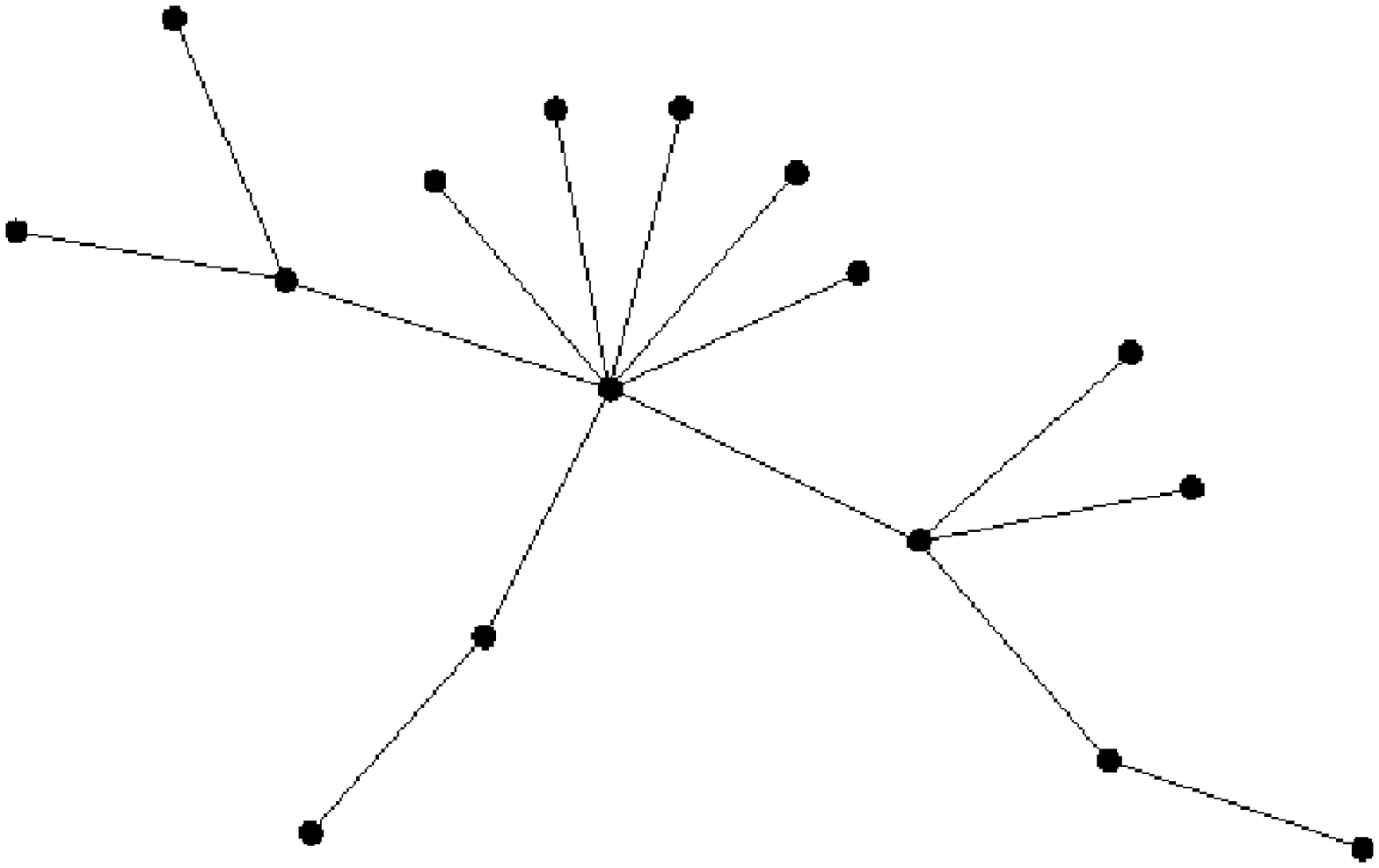}  ~~~
\textbf{Scale-free 2}
\includegraphics[height=.25\textwidth]{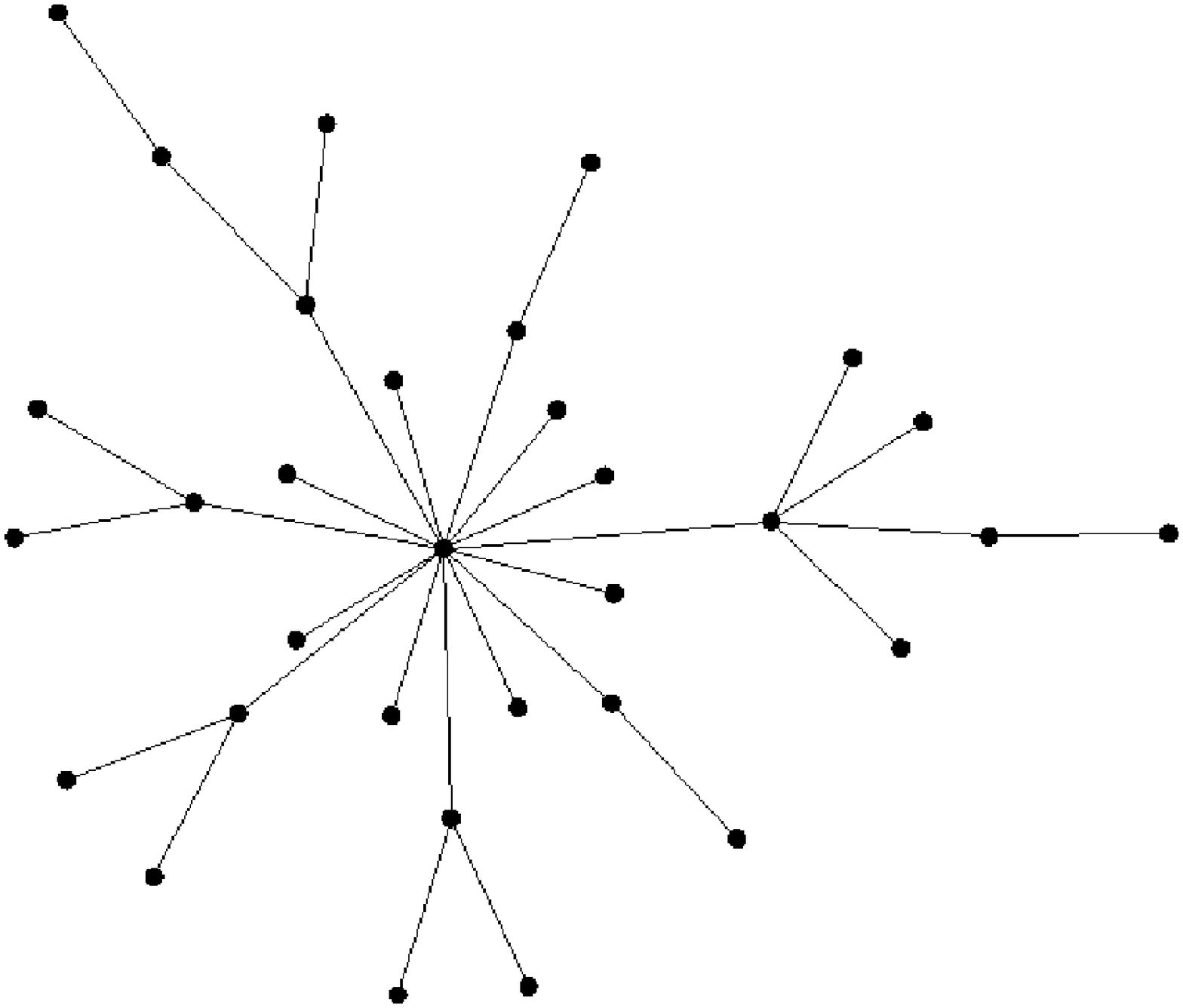}  \\  ~~ \\
\textbf{Scale-free 3}
\includegraphics[height=.25\textwidth]{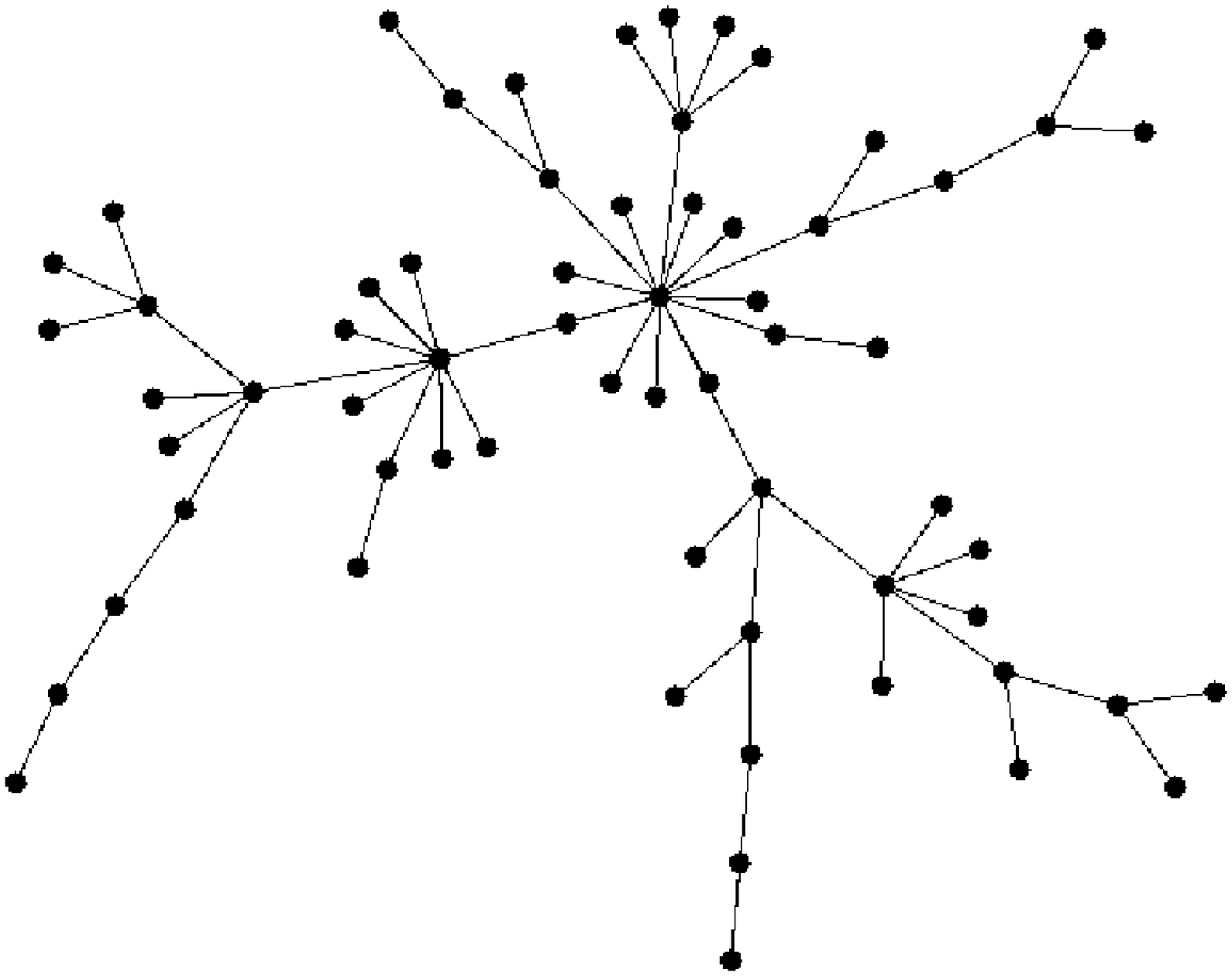}  ~~~
\textbf{Scale-free 4}
\includegraphics[height=.25\textwidth]{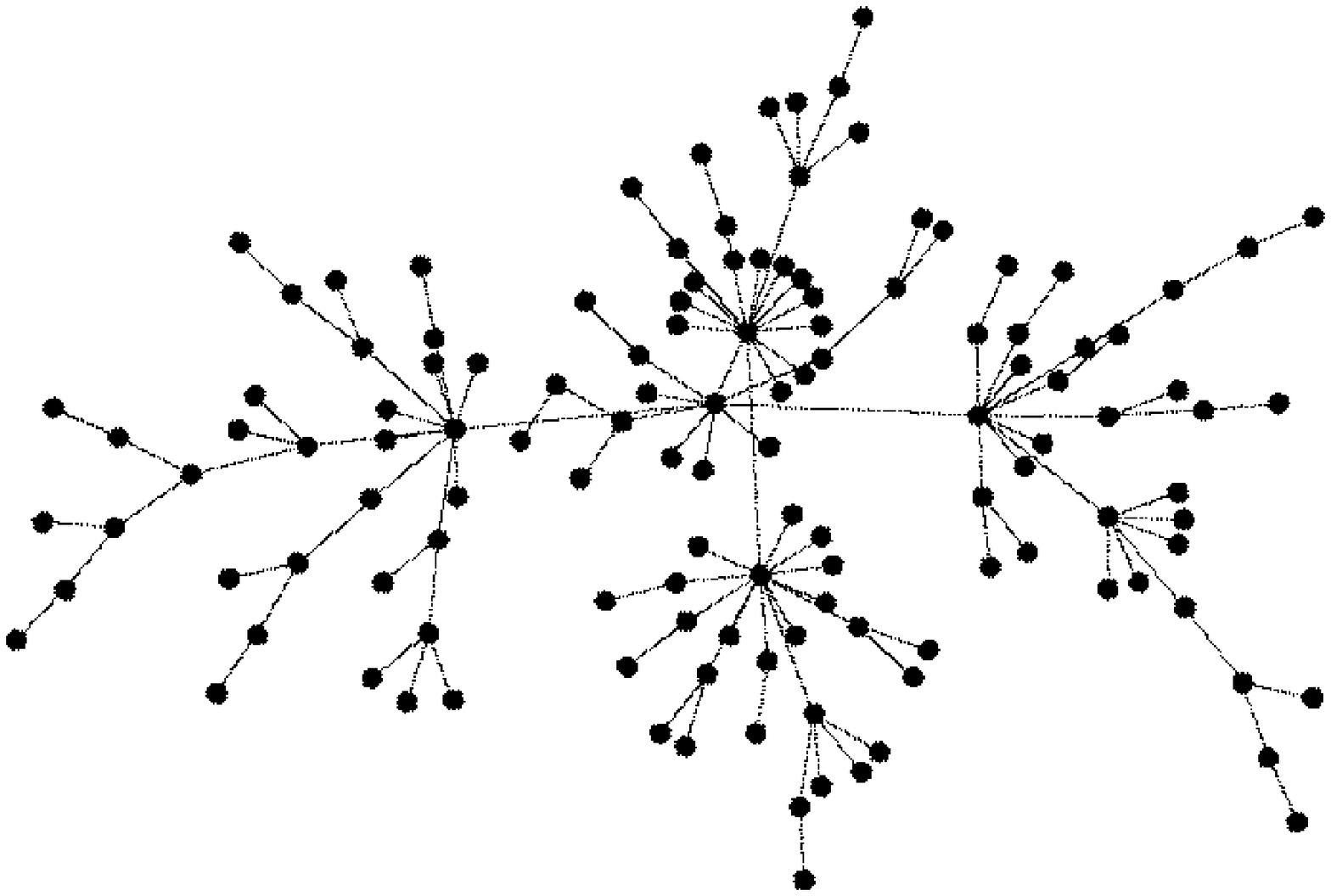} \\  
\caption{Scale-free structures with 16, 32, 64 and 128 nodes  \label{fig:scaleFree}} 
\end{figure}

\begin{figure}[t!]
\scriptsize
% \centering
\textbf{a) Karate}                                            
\includegraphics[height=.35\textwidth,width=.35\textwidth]{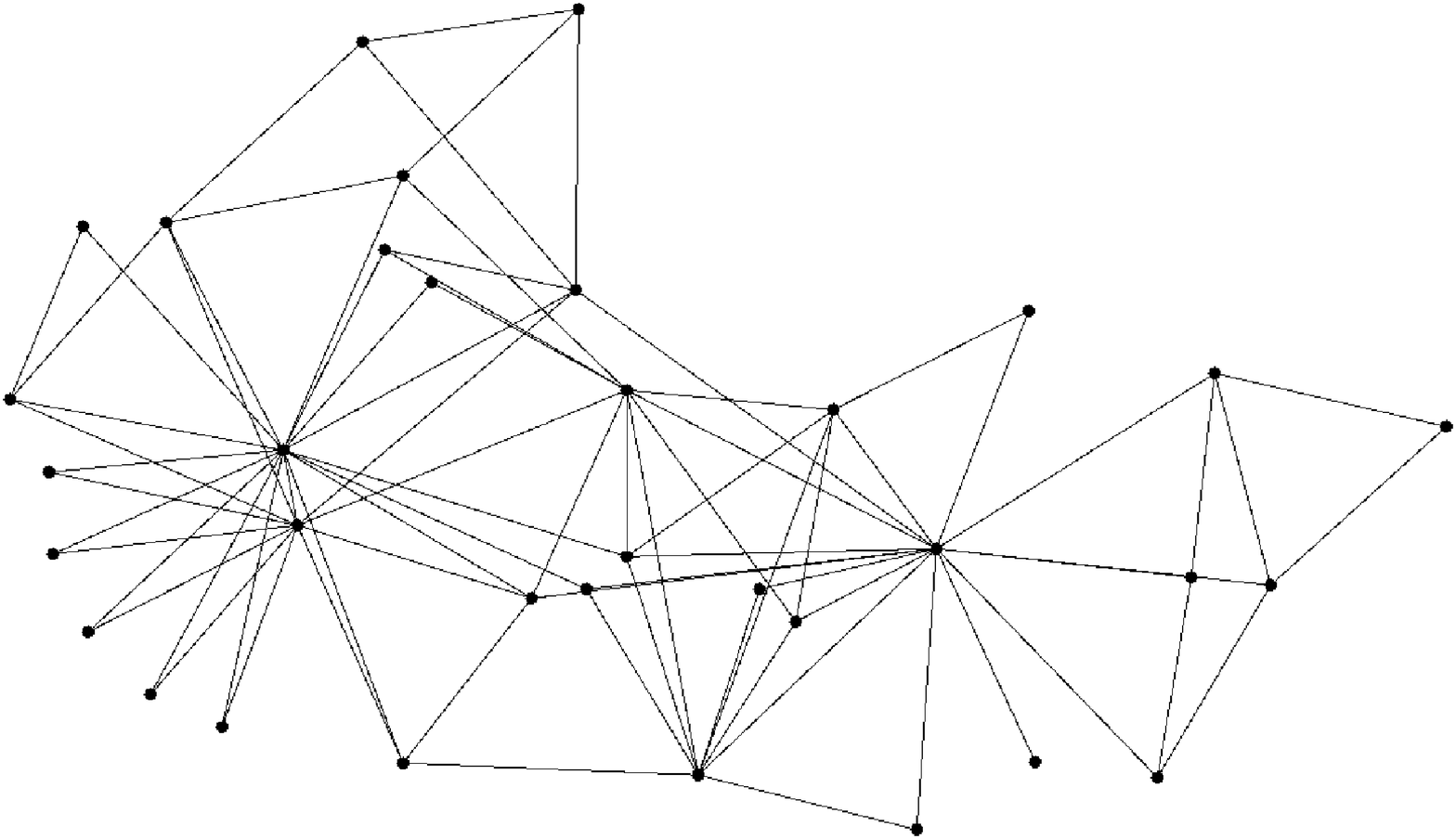} ~~~
\textbf{b) Curtis-54}                                      
\includegraphics[height=.35\textwidth,width=.35\textwidth]{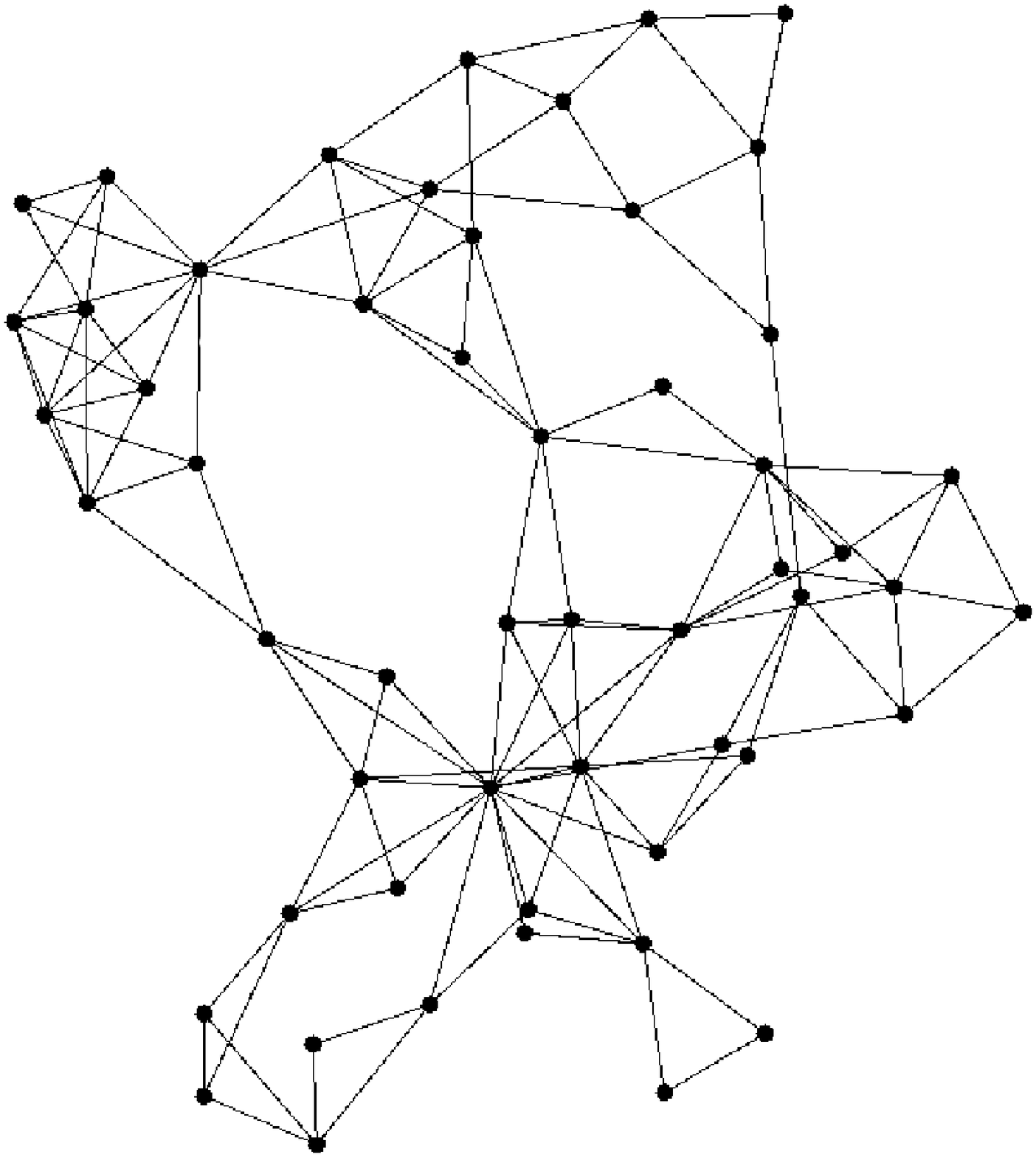}  \\  
\textbf{c) Will-57} 
\includegraphics[height=.35\textwidth,width=.35\textwidth]{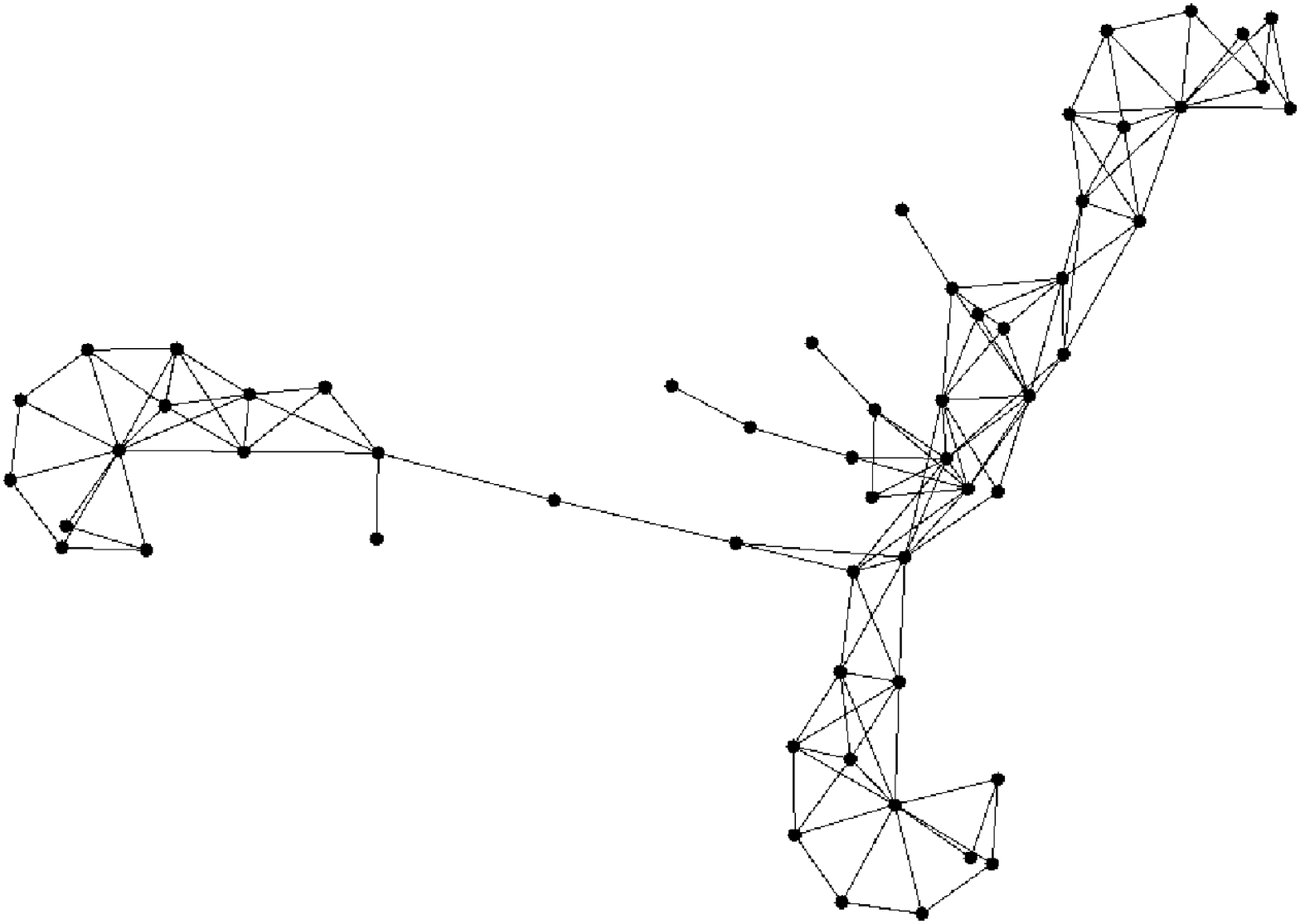}   ~~~
\textbf{d) Dolphins}
\includegraphics[height=.35\textwidth,width=.35\textwidth]{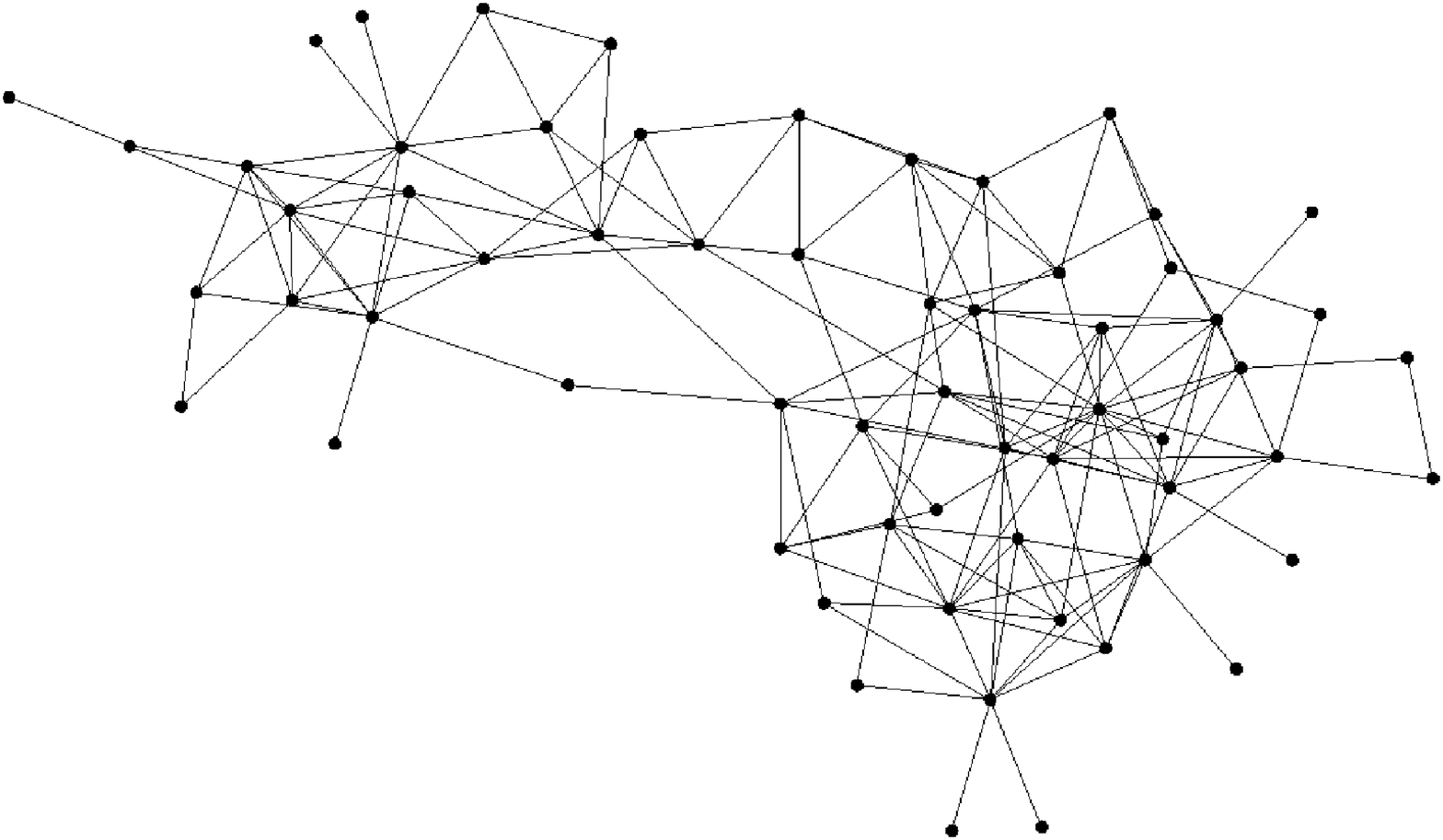} \\  
\textbf{e) Polbooks} 
\includegraphics[height=.35\textwidth,width=.35\textwidth]{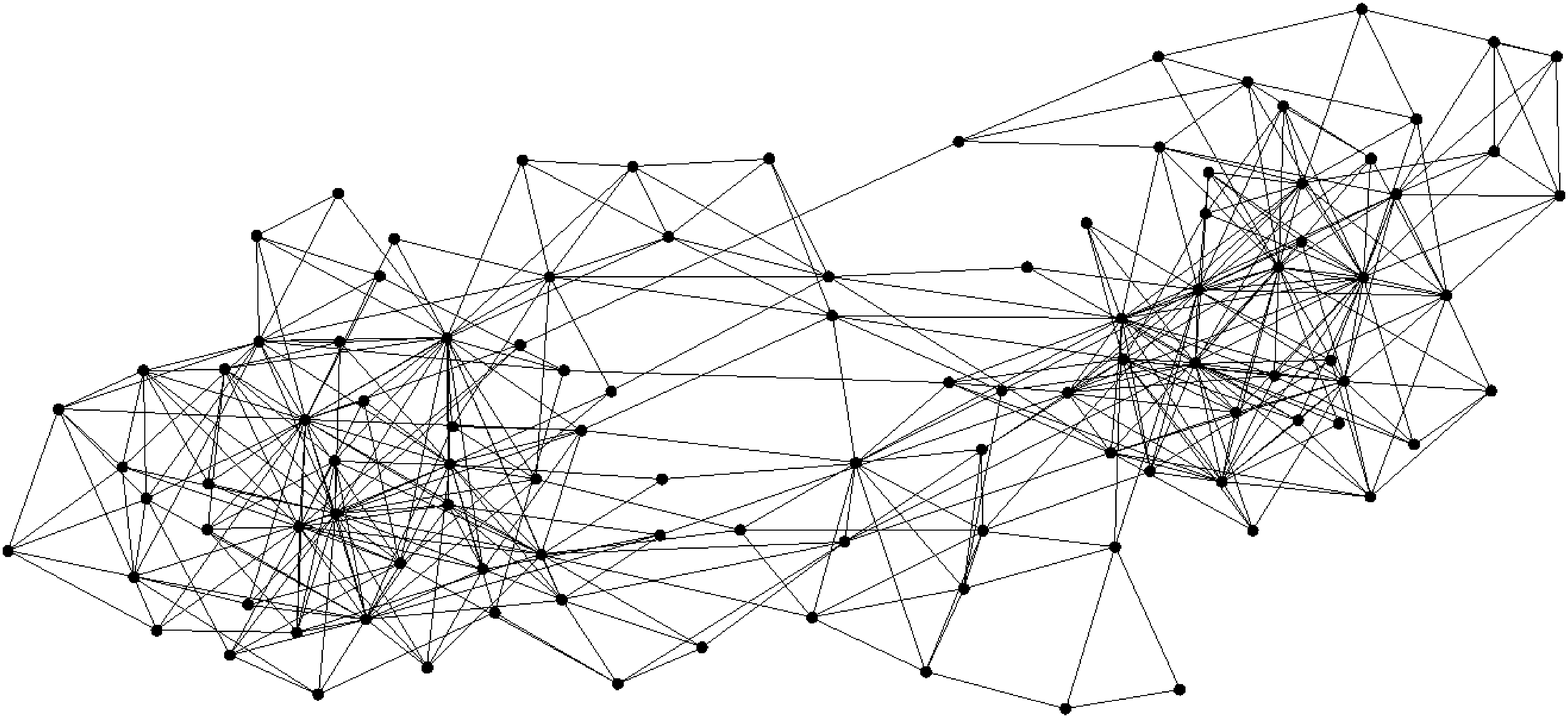}   
\textbf{f) Adj-noun}
\includegraphics[height=.35\textwidth,width=.35\textwidth]{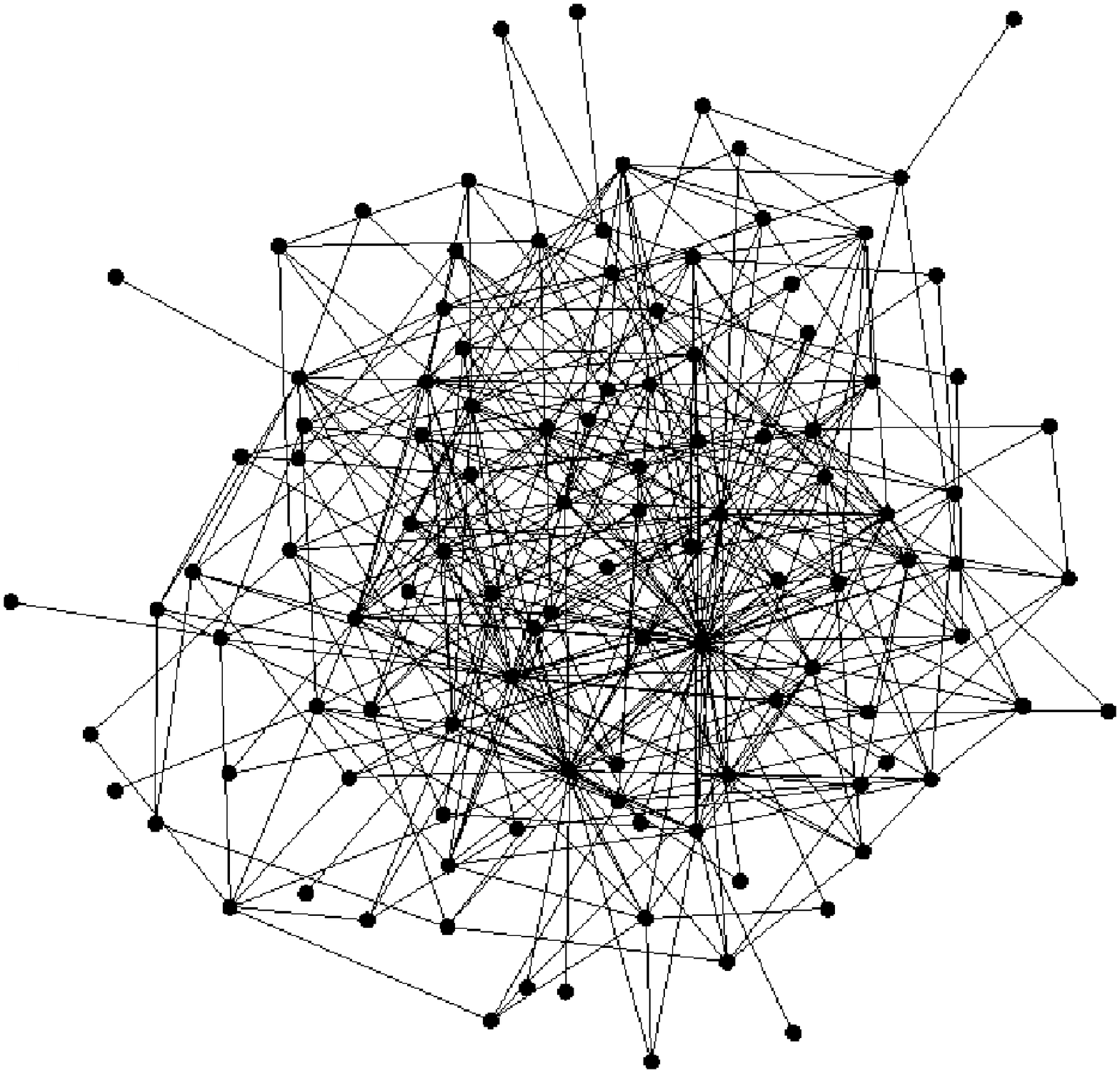} %\\  
% \textbf{g) Arc130}
% \includegraphics[height=.35\textwidth,width=.35\textwidth]{arc130.eps} \\  
\caption{Real-world networks \label{fig:realWorldNets}}  
\end{figure}

The goal in the experiments is to show how the BJP score can improve the quality 
of the structures learned over the competitor scores.
For this, the selected graphs capture the properties of several real-world problems, 
where the target structure has few nodes with large degrees,
and the remaining nodes have very small degree. 
Examples of problems with this characteristic include
gene networks, protein interaction networks and social networks \cite{silva2016machine}.
Thus, for this comparative study, we used three types of structures:
networks with hub topologies, scale-free networks
generated by the Barabasi-Albert model \cite{barabasi2003scale},
and real-world networks, taken from the sparse matrix collection of \cite{davis2011university}.
These structures have an increasing complexity both in $n$ and in $irr$.
The hub networks are shown in Figure~\ref{fig:hubs}, the scale-free networks are shown in 
Figure~\ref{fig:scaleFree}, and the real-world networks are shown in Figure~\ref{fig:realWorldNets}.


% \begin{table}
% \centering
% \begin{tabular}{cccc}
% %  \hline
% 	    \multicolumn{2}{c}{\textbf{Structure}}	&\textbf{$n$}	&\textbf{$irr$}	\\ \hline
% \multirow{4}{*}{Hubs} 		&Hub 1		&	16		& 392			\\
% 				&Hub 2		& 	32		& 	1916		\\
% 				&Hub 3		& 	64		& 	6624		\\
% 				&Hub 4		& 	128		& 	24496		\\ \hline
% \multirow{4}{*}{Scale-free}	&Scale-free 1	&	16		&	364		\\
% 				&Scale-free 2	& 	32		& 	1612		\\
% 				&Scale-free 3	& 	64		& 	6428		\\
% 				&Scale-free 4	& 	128		& 	26188		\\ \hline
% \multirow{4}{*}{Real networks}	&Karate		& 	34		& 	2044		\\
% 				&Curtis-54	&	54		&	3140		\\  
% 				&Will-57	& 	57		& 	4156		\\ 
% 				&Dolphins	& 	62		& 	6480		\\ %\hline
% \end{tabular}
% \caption{Size and irregularity for graphs of Figures \ref{fig:hubs}, \ref{fig:scaleFree} and \ref{fig:realWorldNets}. \label{table:propertieshd}}
% \end{table} 

For each target structure we generated $10$ random distributions 
and $10$ random samples for each distribution, with the Gibbs sampler tool of the Libra toolkit.
Thus, a total of $100$ datasets were obtained for each graph, with the same procedure explained in the previous section.
As a quality measure, we report the average edge Hamming distance between the hundred learned structures 
and the underlying one,
computed as the sum of false positives and false negatives in the learned structure. 
As in the previous section, the algorithms were executed 
for increasing dataset sizes $\mathcal{N}_D=\{250,500,1000,2000,4000,8000\}$,
to assess how their accuracy evolves with data availability. 

% 
% 
% 

% \begin{figure}[t!]
% \scriptsize
% % \centering
% \textbf{a) Karate}                                            
% \includegraphics[height=.35\textwidth,width=.35\textwidth]{karate.png} ~~~
% \textbf{b) Curtis-54}                                      
% \includegraphics[height=.35\textwidth,width=.35\textwidth]{curtis54.png}  \\  
% \textbf{c) Will-57} 
% \includegraphics[height=.35\textwidth,width=.35\textwidth]{will57.png}   ~~~
% \textbf{d) Dolphins}
% \includegraphics[height=.35\textwidth,width=.35\textwidth]{dolphins.png} \\  
% \textbf{e) Polbooks} 
% \includegraphics[height=.35\textwidth,width=.35\textwidth]{polbooks.png}   
% \textbf{f) Adj-noun}
% \includegraphics[height=.35\textwidth,width=.35\textwidth]{adjnoun.png} \\  
% % \textbf{g) Arc130}
% % \includegraphics[height=.35\textwidth,width=.35\textwidth]{arc130.png} \\  
% \caption{Real-world networks \label{fig:realWorldNets2}}  
% \end{figure}
\begin{table*}[!htbp]
\scriptsize
\centering
\scalebox{0.7}{
\begin{tabular}{c|ccc|ccc|ccc}
%  \hline
\textbf{Target} &\textbf{$n$ } & \textbf{$irr$ } & \textbf{$\mathcal{N}_D$ } 	& \multicolumn{3}{c|}{\textbf{Hamming distance}} & \multicolumn{3}{c}{\textbf{Runtime}} \\ %\cline{3-8} 
\textbf{structure} & & &		& \textbf{MPL} & \textbf{IB-score } & \textbf{BJP} & \textbf{MPL} & \textbf{IB-score } & \textbf{BJP} \\ \hline 

\multirow{7}{*}{Hub 1} & \multirow{7}{*}{16} & \multirow{7}{*}{392}
     & 250 & 13.11 (0.07) & 12.36 (0.11) & \textbf{12.14} (0.14)  & 0.16 (0.00) & 0.20 (0.00)  & \textbf{0.06} (0.01)  \\
 & & & 500 & 11.76 (0.06) & 9.92 (0.09) & \textbf{9.42} (0.11)    & 0.14 (0.02) & 0.29 (0.05)  & \textbf{0.11} (0.01)  \\
 & & & 1000 & 10.46 (0.05) & 7.80 (0.11) & \textbf{7.20} (0.12)   & \textbf{0.19} (0.02) & 0.74(0.02)  & 0.25 (0.04)  \\
 & & & 2000 & 9.40 (0.06) & 6.04 (0.11) & \textbf{5.40} (0.11)    & \textbf{0.41} (0.06) & 2.39 (0.08)  & 0.60 (0.07)  \\
 & & & 4000 & 8.19 (0.05) & 4.06 (0.12) & \textbf{3.94} (0.10)    & \textbf{1.09} (0.017) & 6.75 (0.22)  & 1.34 (0.02)  \\
 & & & 8000 & 7.26 (0.05) & 3.16 (0.10) & \textbf{2.88} (0.10)    & 2.908 (0.052) & 17.53 (0.59)  & \textbf{2.59} (0.02)  \\ \hline
%  & 16000 & 6.08 (0.07) & 2.20 (0.10) & 1.96 (0.10) & 22882.40 (419.48) & 41139.90 (1209.48)  & 5250.78 (82.07)  \\
\multirow{7}{*}{Hub 2} & \multirow{7}{*}{32} & \multirow{7}{*}{1916} 
     & 250 & 27.22 (0.12) & 25.73 (0.09) & \textbf{25.02} (0.12)  & 0.42 (4.94) & 0.81 (0.01)  & \textbf{0.39} (0.00)  \\
 & & & 500 & 24.34 (0.11) & 22.00 (0.11) & \textbf{19.98} (0.15)  & \textbf{0.59} (0.00) & 1.50 (0.01)  & 0.92 (0.01)  \\
 & & & 1000 & 21.53 (0.10) & 17.50 (0.12) & \textbf{15.41} (0.16) & \textbf{1.35} (0.02) & 3.87 (0.04)  & 2.15 (0.02)  \\
 & & & 2000 & 18.96 (0.08) & 12.86 (0.13) & \textbf{11.63} (0.11) & \textbf{3.00} (0.05) & 11.39 (0.14)  & 5.28 (0.05)  \\
 & & & 4000 & 16.68 (0.08) & 9.36 (0.12) & \textbf{8.36} (0.11)   & \textbf{7.67} (0.10) & 29.32 (0.36)  & 11.63 (0.09)  \\
 & & & 8000 & 14.56 (0.07) & 7.06 (0.10) & \textbf{6.96} (0.10)   & \textbf{22.45} (0.28) & 76.584 (1.03)  & 23.75 (0.18)  \\ \hline
%  & 16000 & 12.22 (0.08) & 5.04 (0.13) & 5.09 (0.15) & 224934.00 (1790.68) & 369422.00 (6719.26)  & 51312.40 (929.12)  \\
\multirow{7}{*}{Hub 3} & \multirow{7}{*}{64} & \multirow{7}{*}{6624} 
     & 250 & 60.49 (0.21) & 56.55 (0.12) & \textbf{54.03} (0.18)  & 3.09 (0.03) & 1.79 (0.02)  & \textbf{1.37} (0.00)  \\
 & & & 500 & 52.92 (0.19) & 50.60 (0.14) & \textbf{44.88} (0.20)  & 4.90 (63.37) & 4.96 (0.07)  & \textbf{3.86} (0.05)  \\
 & & & 1000 & 46.17 (0.19) & 42.33 (0.19) & \textbf{36.35} (0.25) & \textbf{10.33} (0.11) & 17.24 (0.22)  & 10.39 (0.12)  \\
 & & & 2000 & 40.31 (0.18) & 33.49 (0.24) & \textbf{29.21} (0.29) & \textbf{24.73} (0.28) & 57.95 (0.81)  & 25.991 (0.38)  \\
 & & & 4000 & 34.97 (0.18) & 26.31 (0.25) & \textbf{22.47} (0.30) & \textbf{61.75} (0.66) & 180.92 (3.02)  & 63.64 (0.83)  \\
 & & & 8000 & 30.55 (0.17) & 20.87 (0.29) & \textbf{19.44} (0.31) & 207.48 (2.08) & 627.50 (11.27)  & \textbf{156.24} (3.15)  \\ \hline
%  & 16000 & 26.07 (0.19) & 16.82 (0.36) & 17.77 (0.41) & 1717740.00 (24770.10) & 1907930.00 (32978.50)  & 492457.00 (13900.20)  \\
\multirow{7}{*}{Hub 4} & \multirow{7}{*}{128} & \multirow{7}{*}{24496} 
     & 250 & 134.28 (0.35) & 120.11 (0.14)& \textbf{112.43} (0.28)& 58.92 (0.32) & \textbf{5.86} (0.13)  & 8.31 (0.13)  \\
 & & & 500 & 113.96 (0.28) & 110.03 (0.24)& \textbf{97.25} (0.37) & 78.53 (0.49) & 26.56 (0.42)  & \textbf{25.14} (0.37)  \\
 & & & 1000 & 98.24 (0.29) & 95.01 (0.29) & \textbf{78.39} (0.44) & 129.33 (0.77) & 101.26 (1.05)  & \textbf{74.80} (0.77)  \\
 & & & 2000 & 84.27 (0.26) & 78.78 (0.34) & \textbf{61.35} (0.54) & 259.68 (1.74) & 331.32 (3.19)  & \textbf{198.77} (2.14)  \\
 & & & 4000 & 72.70 (0.23) & 65.17 (0.52) & \textbf{52.11} (0.75) & 777.84 (6.36) & 1252.88 (19.89)  & \textbf{473.05} (6.97)  \\
 & & & 8000 & 62.59 (0.26) & 52.95 (0.78) & \textbf{47.04} (1.03) & 3102.53 (28.43) & 4913.07 (89.81)  & \textbf{1185.91} (23.83)  \\ %\hline
%  & 16000 & 57.31 (0.70) & 45.46 (1.17) & 43.19 (1.15) & 22905400.00 (362220.00) & 16744600.00 (387396.00)  & 4222828.00 (95654.90)  \\

% \multirow{6}{*}{1} & \multirow{6}{*}{\includegraphics[width=1.5cm]{exp/bf/randomParams/graphs/generatingModel3.n6.eps}} 
\end{tabular} }
\caption{Structures with hub topology: 
average and standard deviation of the Hamming distance and runtime (in seconds) 
over 100 repetitions. The best average results are in bold.  \label{table:hubs}}
\end{table*} 

Table~\ref{table:hubs} shows the comparison of 
BJP against MPL and IB-score for the hub structures of Figure~\ref{fig:hubs}.
The table shows the structures, their sizes $n$, and their irregularities, in the first, second and third columns, respectively. 
The dataset sizes $\mathcal{N}_D$ are in the fourth column. 
The fifth column shows the average and standard deviation of the Hamming distance over the $100$ repetitions.
The sixth column shows the corresponding runtimes (in seconds)\footnote{All the experiments were 
performed on an Intel(R) Core(TM) i7-4770 CPU, with 3.40GHz, and 32 GB of main memory.}.
When analyzing these results, it can be seen that for all the algorithms 
the more complex the underlying structure (determined by $n$ and $irr$),
the larger is the number of structural errors for any score and any value of $\mathcal{N}_D$.
The results show that BJP obtains the best performance for all the cases, 
reducing the number of average errors of the structures learned by MPL and IB-score.
It can be seen that, for all the target structures, again MPL has the slowest convergence in $\mathcal{N}_D$.
When compared with both MPL and IB-score, 
the improvements of the BJP score are larger as the complexity
($n$ and $irr$) grows. 
These improvements are statistically significant for all the cases against MPL.
Against IB-score, the improvements of BJP 
are statistically significant for all the cases, except three.
In general, these results confirm that the approximation of BJP is more accurate 
as $n$ and $irr$ grow.
In terms of the respective runtimes, 
the optimization using the BJP score
obtains in general runtimes comparable  to MPL and IB-score.
For the case of Hub 4, BJP shows the best runtime for all the cases where $\mathcal{N}_D > 250$.
This is because the more complex the underlying structure
the better the convergence of the BJP score to correct structures.

% \begin{landscape}
\begin{table*}[!htbp]
\scriptsize
\centering
\scalebox{0.7}{
\begin{tabular}{c|ccc|ccc|ccc}
%  \hline
\textbf{Target} &\textbf{$n$ } & \textbf{$irr$ } & \textbf{$\mathcal{N}_D$ } 	& \multicolumn{3}{c|}{\textbf{Hamming distance}} & \multicolumn{3}{c}{\textbf{Runtime}} \\ %\cline{3-8} 
\textbf{structure} & & &		& \textbf{MPL} & \textbf{IB-score } & \textbf{BJP} & \textbf{MPL} & \textbf{IB-score } & \textbf{BJP} \\ \hline 

\multirow{7}{*}{Scale-free 1} & \multirow{7}{*}{16} & \multirow{7}{*}{364}
     & 250 & 12.35 (0.35) & 11.30 (1.63) & \textbf{11.20} (1.23)   & 0.12 (0.01) & 0.33 (0.11)  & \textbf{0.12} (0.02)  \\ 
 & & & 500 & 10.63 (0.26) & 10.00 (1.45) & 10.00 (1.59)            & \textbf{0.11} (0.01) & 0.40 (0.10)  & 0.16 (0.03)  \\ 
 & & & 1000 & \textbf{9.14} (0.28) & 7.10 (1.64) & 7.30 (1.15)     & \textbf{0.25} (0.02) & 0.76 (0.16)  & 0.35 (0.03)  \\ 
 & & & 2000 & 7.53 (0.23) & \textbf{5.10} (1.07) & 5.20 (1.04)     & \textbf{0.70} (63.75) & 1.88 (0.41)  & 0.71 (0.10)  \\ 
 & & & 4000 & 6.21 (0.22) & 3.70 (0.75) & \textbf{3.50} (0.90)     & 1.90 (0.14) & 3.87 (1.10)  & \textbf{1.41} (0.13)  \\ 
 & & & 8000 & 4.92 (0.22) & 2.30 (1.00) & 2.30 (1.11)              & 6.45 (0.71) & 7.91 (1.85)  & \textbf{2.91} (0.27)  \\ \hline
\multirow{7}{*}{Scale-free 2} & \multirow{7}{*}{32} & \multirow{7}{*}{1612} 
     & 250 & 27.51 (0.45) & 26.50 (1.74) & \textbf{25.88} (2.00)   & \textbf{0.50} (0.03) & 0.92 (0.22)  & 0.56 (0.24)  \\ 
 & & & 500 & 24.08 (0.46) & 22.40 (2.13) & \textbf{20.38} (2.72)   & \textbf{0.78} (0.05) & 1.34 (0.25)  & 0.95 (0.24)  \\ 
 & & & 1000 & 20.82 (0.42) & 18.30 (2.00) & \textbf{17.12} (2.15)  & 2.11 (0.16) & 4.34 (1.15)  & \textbf{1.73} (0.33)  \\ 
 & & & 2000 & 18.27 (0.37) & 13.60 (1.34) & \textbf{12.12} (1.60)  & 5.18 (0.35) & 19.52 (8.80)  & \textbf{5.20} (1.92)  \\ 
 & & & 4000 & 16.13 (0.31) & \textbf{10.40} (1.77) & 10.50 (2.03)  & 12.57 (0.81) & 77.37 (36.80)  & \textbf{10.51} (2.97)  \\ 
 & & & 8000 & 14.41 (0.33) & \textbf{6.56} (1.70) & 7.00 (1.28)    & 41.33 (3.86) & 354.14 (207.98)  & \textbf{25.32} (10.33)  \\ \hline
\multirow{7}{*}{Scale-free 3} & \multirow{7}{*}{64} & \multirow{7}{*}{6428}
     & 250 & 59.11 (0.91) & 57.75 (5.67) & \textbf{55.33} (2.29)   & 4.73 (0.24) & 3.83 (3.35)  & \textbf{2.11} (1.10)  \\ 
 & & & 500 & 50.14 (0.81) & 52.00 (7.01) & \textbf{44.00} (8.64)   & 8.20 (0.45) & 9.85 (4.38)  & \textbf{6.06} (2.92)  \\ 
 & & & 1000 & 43.05 (0.73)& 43.25 (13.98) & \textbf{36.00} (11.04) & 19.54 (1.03) & 29.05 (21.09)  & \textbf{13.87} (6.11)  \\ 
 & & & 2000 & 36.71 (0.74) & 33.50 (9.97) & \textbf{27.67} (8.26)  & 46.99 (2.34) & 95.44 (49.23)  & \textbf{46.06} (9.17)  \\ 
 & & & 4000 & 31.37 (0.56) & 26.25 (4.93) & \textbf{21.33} (2.29)  & 122.24 (6.49) & 275.86 (69.80)  & \textbf{99.06} (18.66)  \\ 
 & & & 8000 & 27.52 (0.57) & 19.00 (3.71) & \textbf{16.00} (7.15)  & 433.09 (22.47) & 1124.33 (841.27)  & \textbf{221.92} (4.05)  \\ \hline
\multirow{7}{*}{Scale-free 4} & \multirow{7}{*}{128} & \multirow{7}{*}{26188}
    &  250 & 131.42 (1.94) & 123.20 (1.09) & \textbf{116.40} (2.73)& 72.69 (3.47) & \textbf{6.71} (1.29)  & 12.50 (4.35)  \\ 
& & &  500 & 109.44 (1.75) & 110.70 (2.96) & \textbf{101.00} (3.85)& 106.37 (5.03) & 36.51 (6.21)  & \textbf{30.74} (12.14)  \\ 
 & & & 1000 & 91.47 (1.58) & 93.00 (4.02) & \textbf{83.10} (4.75)  & 196.18 (9.51) & 140.10 (17.46)  & \textbf{95.14} (31.07)  \\ 
 & & & 2000 & 77.47 (1.40) & 79.20 (5.12) & \textbf{64.50} (5.94)  & 429.12 (24.21 ) & 403.99 (60.94)  & \textbf{271.96} (93.17)  \\ 
 & & & 4000 & 65.44 (1.30) & 62.00 (4.99) & \textbf{46.50} (4.84)  & 1202.84 (92.93) & 1469.52 (283.02)  & \textbf{634.66} (95.49)  \\ 
 & & & 8000 & 57.09 (1.21) & 47.90 (3.87) & \textbf{34.30} (3.92) & 5103.34 (531.26) & 7650.26 (2037.82)  & \textbf{1736.44} (437.66)  \\  %\hline
 \end{tabular}}
\caption{Scale-free networks models: average and standard deviation of the Hamming distance and runtime (in seconds) 
over 100 repetitions. 
The best average results are in bold. \label{table:scaleFree}}
\end{table*} 

Table \ref{table:scaleFree} shows the comparison of 
BJP against MPL and IB-score for the scale-free networks of Figure~\ref{fig:scaleFree}.
The information of the table is organized in the same way as in Table~\ref{table:hubs}.
In contrast with the hub structures, in the scale-free networks
the size of the blankets in the underlying network is more variable.
This can explain the differences in the trends of the Hamming distance, 
when compared with the results obtained for the hub networks.
It can be seen that for all the cases BJP 
reduces the number of average errors of MPL.
The improvements over MPL are statistically significant for all the cases,
except three. 
When compared with IB-score, BJP shows better average number of errors for all the cases, 
except four. Those improvements over IB-score are statistically significant
only for the Scale-free 4 model. 
The best improvements of BJP over MPL can be seen for the Scale-free 4 model, 
$\mathcal{N}_D = 8000$, with improvements of more than $20$ edges corrected.
Against IB-score, the best improvements of BJP can be seen also for the Scale-free 4 model, 
$\mathcal{N}_D = 4000$, with improvements of more than $15$ edges corrected.
In general, these results confirm that the approximation of BJP is more accurate 
as $n$ and $irr$ grow. 

% 
% BPJ the improvements of BJP are larger as the complexity of the underlying network grow ($n$ and $irr$). 
% In contrast with the hub structures, in the scale-free networks
% the size of the blankets is more variable.
% This can explain some diferences in the trends of the Hamming distance, 
% when compared with the results obtained for the hub networks.
% % \textbf{OJO ac\'a porque primero dec\'is que las tendencias son similares y que luego hay diferencias. Aclarar en qu\'e son similares y en qu\'e no. }
% For the two most complex structures (scale-free 3 and 4), 
% BJP reduces the number of errors of the structures learned in all the cases.
% In terms of the respective runtimes, BJP obtains the best runtimes for almost all the cases.
% Specifically, for scale-free 2, for all the cases where $\mathcal{N}_D > 500$;
% for scale-free 3, for all the cases; and 
% for scale-free 4, for all the cases where $\mathcal{N}_D > 250$.
% As the complexity of the target structures grows,
% we can see a better convergence of the BJP score to correct structures. 
% % \textbf{REVISAR porque ser\'ia bueno
% % detallar mejor qu\'e significa "better convergence", es decir, entre cu\'ales casos y/o algoritmos}

\begin{table*}[!htbp]
\scriptsize
\centering
\scalebox{0.7}{
\begin{tabular}{c|ccc|ccc|ccc}
%  \hline
\textbf{Target} &\textbf{$n$ } & \textbf{$irr$ } & \textbf{$\mathcal{N}_D$ } 	& \multicolumn{3}{c|}{\textbf{Hamming distance}} & \multicolumn{3}{c}{\textbf{Runtime}} \\ %\cline{3-8} 
\textbf{structure} & & &		& \textbf{MPL} & \textbf{IB-score } & \textbf{BJP} & \textbf{MPL} & \textbf{IB-score } & \textbf{BJP} \\ \hline 
 \multirow{7}{*}{Karate} & \multirow{7}{*}{34} & \multirow{7}{*}{2044} 
     & 250  & 58.60 (2.78) & 51.91 (3.74)         & \textbf{51.90} (3.59) & 5.30 (1.85)              & 5.01 (1.20)        & \textbf{1.78} (0.24)  \\ 
  & && 500  & 49.80 (2.26) & 44.00 (4.92)         & \textbf{42.20} (3.33) & 12.01 (4.97)            & 14.59 (6.37)        & \textbf{2.95} (0.29)  \\ 
  & && 1000 & 44.00 (2.18) & 27.25 (5.25)         & \textbf{26.00} (4.55) & 22.78 (4.47)            & 213.57 (75.01)      & \textbf{11.07} (3.09) \\ 
  & && 2000 & 40.50 (1.24) & 17.12 (4.89)         & \textbf{11.30} (3.15) & \textbf{40.74} (3.74)   & 1220.47 (656.76)    & 51.54 (10.90)         \\ 
  & && 4000 & 38.00 (0.68) & 7.88 (2.11)          & \textbf{5.80} (1.93)  & \textbf{118.66} (12.99) & 9557.99 (3025.38)   & 195.52 (48.87)        \\ 
  & && 8000 & 36.60 (0.65) & \textbf{2.60} (0.49) & 3.20 (0.82)           & \textbf{320.12} (26.48) & 30963.00 (3032.01)  & 665.70 (89.15)        \\ \hline
\multirow{7}{*}{Curtis-54} & \multirow{7}{*}{54} & \multirow{7}{*}{3140}
     & 250  & 76.50 (1.50)  & 77.00 (2.72)        & \textbf{71.20} (2.37) & 12.09 (0.51)            & 11.64 (1.57)        & \textbf{5.42} (0.34)  \\ 
  & && 500  & 64.40 (1.31)  & 59.10 (2.33)        & \textbf{56.60} (2.00) & 28.82 (1.79)            & 28.49 (3.23)        & \textbf{11.33} (0.42)  \\ 
  & && 1000 & 52.40 (0.86) & 40.10 (1.63)         & \textbf{39.40} (2.15) & 83.15 (3.25)            & 83.29(5.36)         & \textbf{29.48} (1.10)  \\ 
  & && 2000 & 40.10 (0.82) & 22.70 (2.33)         & \textbf{18.90} (3.75) & 244.77 (9.30)           & 278.23 (37.50)      & \textbf{86.36} (6.45)  \\ 
  & && 4000 & 30.30 (1.12) & 7.50 (1.55)          & \textbf{4.40} (1.47)  & 689.46 (26.14)          & 1466.95 (599.15)    & \textbf{240.60} (7.56)  \\ 
  & && 8000 & 24.00 (0.57) & \textbf{2.12} (0.81) & 2.20 (0.64)           & 2015.04 (54.97)         & 4665.29 (788.77)    & \textbf{742.01} (23.51)  \\ \hline
\multirow{7}{*}{Will-57} & \multirow{7}{*}{57} & \multirow{7}{*}{4156} 
     &  250 & 79.50 (1.82) & 81.90 (4.17)         & \textbf{79.40} (4.25) & 13.14 (0.52) & 9.67 (1.38)  & \textbf{5.69} (0.49)  \\ 
  & &&  500 & 66.80 (1.34) & 63.60 (2.27)         & \textbf{60.70} (3.77) & 31.49 (1.93) & 25.19 (2.42)  & \textbf{12.33} (0.92)  \\ 
  & && 1000 & 55.60 (1.08) & 44.50 (2.21)         & \textbf{42.50} (3.78) & 85.83 (3.36) & 75.41 (6.13)  & \textbf{31.91} (2.33)  \\ 
  & && 2000 & 47.60 (0.53) & 25.30 (2.40)         & \textbf{23.80} (2.94) & 232.10 (8.80) & 245.99 (25.06)  & \textbf{87.38} (4.42)  \\ 
  & && 4000 & 38.30 (0.89) & 10.70 (2.03)         & \textbf{9.40} (4.22) & 672.76 (19.39) & 886.78 (123.50)  & \textbf{274.24} (24.75)  \\ 
  & && 8000 & 28.90 (0.54) & \textbf{2.70} (0.53) & 3.90 (1.75) & 2383.00 (72.06) & 3077.88 (418.76)  & \textbf{787.45} (53.24)  \\ \hline
\multirow{7}{*}{Dolphins} & \multirow{7}{*}{62} & \multirow{7}{*}{6480} & 250 & 126.70 (4.00) & 126.90 (5.18) & \textbf{125.20} (4.14) & 24.02 (2.64) & 12.46 (3.02)  & \textbf{7.11} (1.35)  \\ 
  && & 500 & 106.60 (4.32) & 106.10 (6.06) & \textbf{102.10} (5.10) & 48.47 (5.52) & 30.53 (5.83)  & \textbf{16.73} (2.34)  \\ 
  & && 1000 & 88.50 (1.90) & 71.60 (4.64) & \textbf{65.90} (3.84) & 126.57 (12.52) & 120.44 (13.49)  & \textbf{55.37} (3.75)  \\ 
  & && 2000 & 74.20 (2.02) & 50.60 (3.41) & \textbf{47.30} (3.52) & 349.26 (23.90) & 337.56 (26.13)  & \textbf{144.14} (12.24)  \\ 
  & && 4000 & 63.00 (1.99) & 32.50 (2.93) & \textbf{27.70} (3.14) & 981.07 (65.15) & 1092.66 (102.50)  & \textbf{386.27} (25.09)  \\ 
  & && 8000 & 50.80 (1.60) & 20.60 (1.94) & \textbf{12.90} (2.06) & 3591.12 (153.37) & 4171.51 (173.19)  & \textbf{1331.72} (44.67)  \\\hline
\multirow{7}{*}{Polbooks} & \multirow{7}{*}{105} & \multirow{7}{*}{30374} 
      & 250 & 513.53 (5.94) & 435.06 (1.19) & \textbf{428.53} (1.96)           & 48046.00 (18543.70) & 4.33 (0.67)  & \textbf{4.1} (0.58)  \\
  & &&  500 & 479.00 (6.30) & 428.28 (3.07) & \textbf{418.27} (3.96)  & 48046.00 (3116.25)    & 13.48 (2.49)  & \textbf{10.89} (2.17)  \\
  & &&  1000 & 439.00 (23.10) & 414.11 (4.11) & \textbf{407.93} (4.05)& 8532.64 (7682.98)      & 56.37 (8.57)  & \textbf{29.30} (3.81)  \\
  &  && 2000 & 409.43 (7.92) & 399.72 (4.38) & \textbf{393.53} (4.68) & 1455.56 (203.59)       & 170.71 (18.60)  & \textbf{85.33} (9.02)  \\
  & &&  4000 & 378.86 (7.12) & 381.72 (6.04) & \textbf{375.60} (6.41) & 2344.57 (722.87)      & 648.49 (190.16)  & \textbf{246.44} (27.90)  \\
  & &&  8000 & \textbf{353.14} (4.91) & 364.07 (7.17) & 357.50 (5.30) & 4462.01 (1383.34)      & 3726.04 (1105.15)  & \textbf{971.54} (168.34)  \\\hline
\multirow{7}{*}{Adj-Noun} & \multirow{7}{*}{112} & \multirow{7}{*}{39728}   & 250 & 505.60 (9.13) & 424.30 (0.58) & \textbf{422.10} (2.96) & 30664.20 (22276.90) & \textbf{1.86} (0.45)  & 4.49 (2.12)  \\
  & && 500 & 461.30 (9.60) & 421.20 (1.84) & \textbf{412.90} (2.75) & 2146.86 (3246.57) & \textbf{6.56} (2.22)  & 9.38 (2.32)  \\
  & &&  1000 & 430.90 (6.28) & 413.80 (2.35) & \textbf{401.60} (3.26) & 521.98 (63.52) & 27.30 (5.91)  & \textbf{27.00} (5.23)  \\
  & &&  2000 & 399.70 (8.45) & 400.80 (3.61) & \textbf{387.50} (5.23) & 814.73 (207.03) & 108.95 (17.35)  & \textbf{80.53} (13.08)  \\
  & &&  4000 & \textbf{372.00} (4.61) & 384.40 (6.23) & 373.80 (5.60) & 1465.95 (275.36) & 449.33 (118.64)  & \textbf{224.86} (31.93)  \\
  & &&  8000 & \textbf{347.30} (6.99) & 366.30 (5.90) & 356.30 (4.92) & 3443.87 (816.50) & 2172.69 (749.05)  & \textbf{807.17} (243.59)  \\\hline
% 
% \multirow{7}{*}{arc130} & \multirow{7}{*}{130} & \multirow{7}{*}{90020} & 250 & 714.00 (0.00) & 715.40 (0.36) & 721.60 (1.97) & 61.50 (19.26) & 2729.50 (568.42)  & 9155.80 (5741.45)  \\
%   && & 500 & 714.00 (0.00) & 715.50 (1.01) & 704.90 (7.86) & 68.00 (17.43) & 7185.00 (4547.19)  & 252124.00 (318215.00)  \\
%  & &&  1000 & 714.00 (0.00) & 714.80 (1.23) & 689.80 (4.42) & 88.90 (35.16) & 11497.50 (6900.63)  & 4263730.00 (7851240.00)  \\
%  & &&  2000 & 720.57 (8.47) & 711.50 (3.39) & 691.40 (7.77) & 102.00 (25.18) & 69593.00 (49411.40)  & 2113780.00 (2609140.00)  \\
%  & &&  4000 & 714.30 (4.64) & 711.20 (3.71) & 688.30 (8.68) & 243.70 (95.39) & 112907.00 (89348.40)  & 1044230.00 (1279530.00)  \\
%  & &&  8000 & 710.97 (3.29) & 711.40 (3.54) & 693.70 (6.14) & 577.70 (194.09) & 283090.00 (227138.00)  & 412507.00 (218032.00)  \\\hline
  
 \end{tabular}}
\caption{Real networks: average and standard deviation of the Hamming distance and runtime (in seconds) 
over 100 repetitions. The best average results are in bold.  \label{table:realNets}}
\end{table*}

Finally, Table \ref{table:realNets} show the results for the real-world networks of Figure~\ref{fig:realWorldNets}.
Again, the information of this table is organized in the same way as in the previous tables. 
The real network structures are ordered by their complexity (in $n$ and $irr$).
The trends in these results are consistent to those in the previous tables. 
For all the networks, BJP improves the average quality of the structures learned for all the cases when $\mathcal{N}_D < 4000$.
When compared with MPL, BJP shows improvements for all the cases, 
except three.  
The best improvements of BJP over MPL can be seen for the Polbooks and Adj-noun networks,  $\mathcal{N}_D = 250$,
with improvements of more than $80$ edges corrected.
This is coherent, since those are the most complex networks, and the best improvements are obtained when data is scarcer.
When compared with IB-score, BJP shows improvements for all the cases 
except three,
with no statistically significant differences. 
The best improvements of BJP over IB-score can be seen for the more complex networks, Polbooks and Adj-noun,  
with improvements of more than $10$ edges corrected.
Regarding the runtimes, it can be seen again that BJP tends to improve the runtime 
over MPL and IB-score for almost all the cases.

In general, the results discussed confirm that BJP always outperforms 
the competitors when data are scarce. 
Also, the improvements are greater both in quality and runtime, for the more complex models. 
This confirms the hypothesis that the approximation proposed by BJP 
can improve the quality of the learning process.

% 
% BJP obtains reduces the number of errors of the structures learned for all the cases.
% In terms of the respective runtimes, again 
% the optimization using BJP tends to improve the runtime 
% over MPL and IBMAP as well as the complexity of the target structures grows.
% 
% Additionally, Table \ref{table:realNets} shows the results of the same experiment 
% for the real networks shown in Figure~\ref{fig:realWorldNets}. 
% The trends in these results are similar than those in the previous tables, in terms of quality and runtime. 
% 
% In terms of the respective runtimes of the three tables, 
% the fourth column shows the corresponding runtimes.
% Clearly, the optimization using the BJP score
% obtains in general better runtimes than using MPL and IBMAP.

% \multirow{4}{*}{Real networks}	&Karate		& 	34		& 	2044		\\
% 				&Curtis-54	&	54		&	3140		\\  
% 				&Will-57	& 	57		& 	4156		\\ 
% 				&Dolphins	& 	62		& 	6480		\\ %\hline

% \include{benchmarkResults}

 \section{Conclusions \label{sec:conclusions}}

In this work we have introduced a novel scoring function for learning the structure of Markov networks. 
The BJP score computes the posterior probability of independence structures
by considering the joint probability distribution of the collection of Markov blankets 
of the structures. The score computes the posterior of each Markov blanket progressively, using
information from other blankets as evidence.
The blanket posteriors of variables with fewer neighbors is computed first,
and then this information is used as evidence for computing the posteriors for variables with bigger blankets.
Thus, BJP can be useful to improve the data efficiency for problems with complex networks,
where the topology exhibits irregularities, such as social and biological networks.
In the experiments, BJP scoring proved that can improve the sample complexity 
of the state-of-the-art competitors. 
The score is tested by using exhaustive search for low-dimensional problems
and by using a heuristic hill-climbing mechanism for higher-dimensional problems.
The results show that BJP produces more accurate structures than the state-of-the-art competitors. 
% Although the runtime of the search mechanism used with BJP 
% compares favorably to the other similar search-and-score solutions (MPL-IAMB-HC and IBMAP-HC),
% it is required to design more effective optimization methods
% to compete with the independence-based algorithms efficiency. 

We will guide our future work toward the design of more effective optimization methods,
since the hill-climbing optimization has two inherent disadvantages: 
i) by only flipping 
one edge per step it scales slowly with the number of variables of the domain $n$, 
ii) it is prone to getting stuck in local optima.
% However, the thesis of this work is only based on the scoring function, 
% instead of its optimization. 
% Also, it is clearly worthwhile considering testing 
% our approach in more practical real-world testbeds, 
% potentially comparing the inference performance of the structures learned,
% against those learned by state-of-the-art score-based algorithms.
Moreover, we consider that the properties of BJP score have considerable potential 
for both further theoretical development, and applications.

\section{Acknowledgements}

This work was supported by Consejo Nacional de Investigaciones Cient\'ificas y T\'ecnicas (CONICET) [PIP 2013 117], 
Universidad Nacional del Litoral (UNL) [CAI+D 2011 548] 
and Agencia Nacional de Promoci\'on Cient\'ifica y Tecnol\'ogica (ANPCyT) [PICT 2014 2627] and [PICT-2012-2731].
% Authors would like to thank four anonymous reviewers for their helpful comments.

% \include{appendix.bayesiantest}

\appendix
\appendix\normalsize
% \clearpage
\begin{appendices}
\section{ - Correctness of BJP \label{app:correctness}}
Based on the developments in Section \ref{sec:bjp}, and the analysis in Section \ref{sec:experiments},
we see that the BJP score is a good measure of the fit of the estimated MN to the dataset.
In this appendix we are concerned about the correctness of the 
method used by BJP to compute the posterior of structures. Thus, by “correctness” we mean that 
the probability computed by BJP is equivalent to the posterior probability of a MN structure. 
% This allows the score to be consistent, that is, 
% as the dataset size tends to infinity
% the maximization of the score returns the true structure. 

In the formulation of the BJP score, the joint distribution of the blankets of $G$ 
is calculated by computing the probabilities of conditional independence and dependence assertions
contained in the blanket of each variable of the domain. 
Our discussion in this appendix follows by demonstrating 
that all the members and non-members of each blanket  
are unequivocally determined in (\ref{eq:blanketposterior}),
and therefore, that the joint posterior over these dependences and independences is equivalent to the posterior of the blankets.
% by the probabilities of independence and dependence computed in the score. 
% 
From \cite[Definition~2]{schluter2014ibmap},
the \textit{Markov blanket closure} is a set of independence and dependence assertions 
that are formally proven to correctly determine a MN structure.
This set is obtained by determining the blanket of each variable $X_i \in X$ 
with the following set of conditional independence and dependence assertions: 
%    \begin{eqnarray} \label{eq:closureMBX}
%    &\Big\{& \ci{X_i}{X_j}{\blanket{i} }  ~:~ X_j \!\notin\! \blanket{i}  \Big\} ~~~ \bigcup \nonumber \\
%    &\Big\{& \cd{X_i}{X_j}{\blanket{i} \!\setminus\! \{X_j\}}  ~:~  X_j \!\in\! \blanket{i}  \Big\}.
%    \end{eqnarray}
   \begin{equation} \label{eq:closureMBX}
   \Big\{ \ci{X_i}{X_j}{\blanket{X_{i}} }  ~:~ X_j \!\notin\! \blanket{X_{i}}  \Big\} ~~~ \bigcup ~~~
   \Big\{ \cd{X_i}{X_j}{\blanket{X_{i}} \!\setminus\! \{X_j\}}  ~:~  X_j \!\in\! \blanket{X_{i}}  \Big\}.
   \end{equation}
Clearly, this is exactly the same set used by BJP in (\ref{eq:blanketposterior})
to compute the posterior of the blanket of each variable of the domain.
Since this set determines all members and non-members of a blanket,
the posterior of this set of assertions is equivalent
to the posterior of the blanket. 
Then, we demonstrate that such probabilities
are correctly estimated by (\ref{eq:ciPosterior}) and (\ref{eq:cdPosterior}).
We proceed by discussing their correctness separately for independence and dependence assertions. 

Equation (\ref{eq:ciPosterior}) computes the probability of independence between a variable
and a non-adjacent variable, conditioned on its blanket,
given the previously computed blankets and the dataset $D$.
In this equation, for the case when $i < k$, 
which indexes over the variables for which the blanket posterior is not already computed,
the posterior of the independence assertion $\ci{\psi_{i}}{\psi_{k}}{\blanket{\psi_{i}}}$ must be computed from data.
It is performed by using the Bayesian statistical test of \cite{MARGARITIS05},
that has been proven to be statistically consistent,
since its mean square error tends to $0$ as the dataset size tends to infinity.
For the case when $i > k$, which indexes over the variables for which the blanket posterior is already computed,
the independence assertion is inferred as $1$, 
since its independence is determined by the blanket of $\psi_k$, which is in the evidence $\left\{ \blanket{\psi_j} \right\}_{j=0}^{i-1}$.
By definition in (\ref{eq:blanketposterior}), 
this case applies to all the variables $\psi_k \notin \blanket{\psi_{i}}$ 
(i.e., all the variables that are not connected to $\psi_i$).
We argue the correctness for this inference by considering an intuitive equivalence 
commonly used by constraint-based approaches
to perform independence tests that involve smaller number of variables
\cite[p.~980]{koller09}.
If two variables $X_i$ and $X_k$ are not neighbors in $G$, 
then by applying the local Markov property of (\ref{eq:lmp})
once for each, we have that $\ci{X_i}{X_k}{\blanket{X_i}}$ and
$\ci{X_i}{X_k}{\blanket{X_k}}$ hold. 
Therefore, the inference made is correct.
% ESTO ESTÁ EXTRAIDO DE PAGINA 980, DE KOLLER, DONDE SE JUSTIFICA COMO USAR MARKOV BLANKETS
% EN ALGORITMOS DE APRENDIZAJE BASADOS EN TESTS DE INDEPENDENCIA

A similar argument can be given for the case of the dependence assertions. 
Equation (\ref{eq:cdPosterior}) computes the probability of dependence between a variable
and an adjacent variable conditioned on its remaining neighbors,
given the previously computed blankets and the dataset $D$.
Again, for the case when $i < k$, 
which indexes over the variables for which the blanket posterior is not already computed,
the posterior of the dependence assertion must be computed from data. 
For the case when $i > k$, which indexes over the variables for which the blanket posterior is already computed,
the dependence assertion is inferred as $1$, 
since its dependence is determined by the blanket of $\psi_k$, 
which is again in the evidence $\left\{ \blanket{\psi_j} \right\}_{j=0}^{i-1}$.
% The proof of the correctness for this inference is similar to the proof given for the independence case. 
By definition in (\ref{eq:blanketposterior}), 
this case applies to all the variables $\psi_k \in \blanket{\psi_{i}}$ (i.e., 
all the variables that are connected to $\psi_i$).
Clearly, if two variables $X_i$ and $X_k$ are neighbors in $G$,
there are no sets separating them in the graph.
Therefore, the dependence assertion inferred is true. 

% As a corollary, it can be seen that the posterior probabilities computed in BJP to approximate
% the posterior of a structure is a subset of the assertions contained in the Markov blanket closure. 
% Given that several probabilities are inferred, the structure is correctly determined by BJP,
% avoiding the computation of potentially expensive and unreliable probabilities.
% Since the probabilities that are computed from data use a statistically consistent method (i.e., the 
% Bayesian statistical test), the BJP score also turns out to be also statistically consistent in the limit of infinite data.

\section{Example of BJP score computation \label{appendix:example}}
 This appendix shows a complete example of the computation of the BJP score for 
 the graph of Figure~\ref{fig:hub}. Consider this graph as the independence structure
 of a probability distribution $\Pr(V)$, with $n=4$ variables
$V=\{X_0,X_1,X_2,X_3\}$, represented by a MN.
Given a dataset $D$, the BJP score can be computed by
following the next steps:

 a) Build the vector $\psi$, with the nodes sorted by their degree in ascending order: 
  $\psi = (X_1,X_2,X_3,X_0)$.  
  
% \end{center}
 b) By following \eq{eq:jointBlankets2}, the computation of $BJP(G)$ is given by:
 \begin{small}
    \begin{eqnarray*} \scriptsize
&BJP(G)=& ~~\Pr \left( \blanket{X_1} \bigg| D \right) \\
&& \times \Pr \left( \blanket{X_2} \bigg| \blanket{X_1}, D \right) \\ 
&& \times \Pr \left( \blanket{X_3} \bigg| \blanket{X_1}, \blanket{X_2}, D \right) \\ 
&& \times \Pr \left( \blanket{X_0} \bigg| \blanket{X_1}, \blanket{X_2}, \blanket{X_3},D \right) .
   \end{eqnarray*}
\end{small}   
 
 c) Compute each term of the above expression by following \eq{eq:blanketposterior}, resulting in:
 \begin{small} 
    \begin{equation*} 
    \begin{aligned}
  \Pr \left( \blanket{X_1} \bigg| D \right) = & ~~\Pr \left( \ci{X_1}{X_2}{X_0} \bigg| D \right)  \\
   &\times \Pr \left( \ci{X_1}{X_3}{X_0} \bigg| D \right) \\
   &\times  \Pr \left( \cd{X_1}{X_0}{\emptyset} \bigg| D \right).\\
 \Pr \left( \blanket{X_2} \bigg| \blanket{X_1}, D \right) = & ~~
  \Pr \left( \ci{X_2}{X_1}{X_0} \bigg| \blanket{X_1},D \right) \\
   & \times \Pr \left( \ci{X_2}{X_3}{X_0} \bigg| \blanket{X_1},D \right) \\ 
   & \times \Pr \left( \cd{X_2}{X_0}{\emptyset}  \bigg| \blanket{X_1},D \right). \\
  \Pr \left( \blanket{X_3} \bigg| \blanket{X_1}, \blanket{X_2}, D \right) = & ~~
  \Pr \left( \ci{X_3}{X_1}{X_0} \bigg| \blanket{X_1},\blanket{X_2},D \right) \\
   &\times \Pr \left( \ci{X_3}{X_2}{X_0} \bigg| \blanket{X_1},\blanket{X_2},D \right) \\
   & \times \Pr \left( \cd{X_3}{X_0}{\emptyset} \bigg| \blanket{X_1},\blanket{X_2},D \right). \\
  \Pr \left( \blanket{X_0} \bigg| \blanket{X_1}, \blanket{X_2}, \blanket{X_3},D \right) = & ~~
  \Pr \left( \cd{X_0}{X_1}{X_2,X_3} \bigg| \blanket{X_1},\blanket{X_2},\blanket{X_3},D \right) \\
   &\times \Pr \left( \cd{X_0}{X_2}{X_1,X_3} \bigg| \blanket{X_1},\blanket{X_2},\blanket{X_3},D \right) \\
   & \times \Pr \left( \cd{X_0}{X_3}{X_1,X_2}\bigg| \blanket{X_1},\blanket{X_2},\blanket{X_3},D \right). \\
    \end{aligned}
    \end{equation*}
 \end{small}

 d) By replacing Equations~(\ref{eq:ciPosterior}) and (\ref{eq:cdPosterior}) 
 in the factors of the above expression, 
 one half of the tests can be inferred, and only the following probabilities must be computed from data by using the Bayesian statistical test: 
 \begin{small} 
    \begin{equation*} 
    \begin{aligned}
  \Pr \left( \blanket{X_1} \bigg| D \right) = \Pr \left( \ci{X_1}{X_2}{X_0} \bigg| D \right) 
    \times \Pr \left( \ci{X_1}{X_3}{X_0} \bigg| D \right) 
    \times \Pr \left( \cd{X_1}{X_0}{\emptyset} \bigg| D \right). \\
  \Pr \left( \blanket{X_2} \bigg| \blanket{X_1}, D \right) = 1 \times \Pr \left( \ci{X_2}{X_3}{X_0} \bigg| D \right) 
   \times \Pr \left( \cd{X_2}{X_0}{\emptyset}  \bigg| D \right). \\
  \Pr \left( \blanket{X_3} \bigg| \blanket{X_1}, \blanket{X_2}, D \right) = 
  1 \times 1 \times  \Pr \left( \cd{X_3}{X_0}{\emptyset} \bigg| D \right). \\
\end{aligned}
  \end{equation*}
  \end{small}
  \begin{small} 
    \begin{equation*} 
    \begin{aligned}
  &\Pr \left( \blanket{X_0} \bigg| \blanket{X_1}, \blanket{X_2}, \blanket{X_3},D \right) = 1 \times 1 \times 1.
\end{aligned}
   \end{equation*} 
\end{small} The inferred tests are the 1s at each equation.       

\end{appendices}

% \section*{References}
% \bibliography{mybibfile}
% \bibliography{references}

 \newcommand{\noop}[1]{}

\end{document}